\PassOptionsToPackage{round,authoryear}{natbib}
\PassOptionsToPackage{table,dvipsnames}{xcolor}
\documentclass[]{style/company_light} % switch to style/company_border for border variant
\usepackage[utf8]{inputenc}
\usepackage{hyperref}
\usepackage{url}
\usepackage{booktabs}
\usepackage{amsfonts}
\usepackage{amsmath}
\usepackage{nicefrac}
\usepackage{microtype}
\usepackage{graphicx}
\usepackage{subcaption}
\usepackage{multirow}
\usepackage{enumerate}
\usepackage{enumitem}
\usepackage[most]{tcolorbox}
\usepackage{adjustbox}
\usepackage{array}
\usepackage{amssymb}

\newcommand{\yes}{\textcolor{ForestGreen}{$\checkmark$}}
\newcommand{\no}{\textcolor{red}{$\times$}}
\newcommand{\partialyes}{\textcolor{orange}{$\checkmark$}}

\graphicspath{{imgs/}}

\title{Frontier-Eng: Benchmarking Self-Evolving Agents on Real-World Engineering Tasks with Generative Optimization}
\author{Navers Lab, Einsia.AI}
\abstract{%
Current LLM agent benchmarks, which predominantly focus on binary pass/fail tasks such as code generation or search-based question answering, often neglect the value of real-world engineering that is often captured through the iterative optimization of feasible designs. To this end, we introduce \textbf{Frontier-Eng}, a human-verified benchmark for \emph{generative optimization}---an iterative propose--execute--evaluate loop in which an agent generates candidate artifacts, receives executable verifier feedback, and revises them under a fixed interaction budget---spanning $47$ tasks across five broad engineering categories. Unlike previous suites, Frontier-Eng tasks are grounded in industrial-grade simulators and verifiers that provide continuous reward signals and enforce hard feasibility constraints under constrained budgets. We evaluate eight frontier language models using representative search frameworks, finding that while \texttt{GPT-5.4} achieves the most robust performance, the benchmark remains challenging for all models. Our analysis suggests a dual power-law decay in improvement frequency ($\sim$ 1/iteration) and magnitude ($\sim$ 1/improvement count). We further show that although width improves parallelism and diversity, depth remains crucial for hard-won improvements under a fixed budget. Frontier-Eng establishes a new standard for assessing the capacity of AI agents to integrate domain knowledge with executable feedback to solve complex, open-ended engineering problems.

}

\date{\today}
\metadata[HomePage]{\url{https://lab.einsia.ai/frontier-eng}}

\begin{document}

\maketitle

\section{Introduction}

\begin{figure*}[t]
\centering
\includegraphics[width=\linewidth]{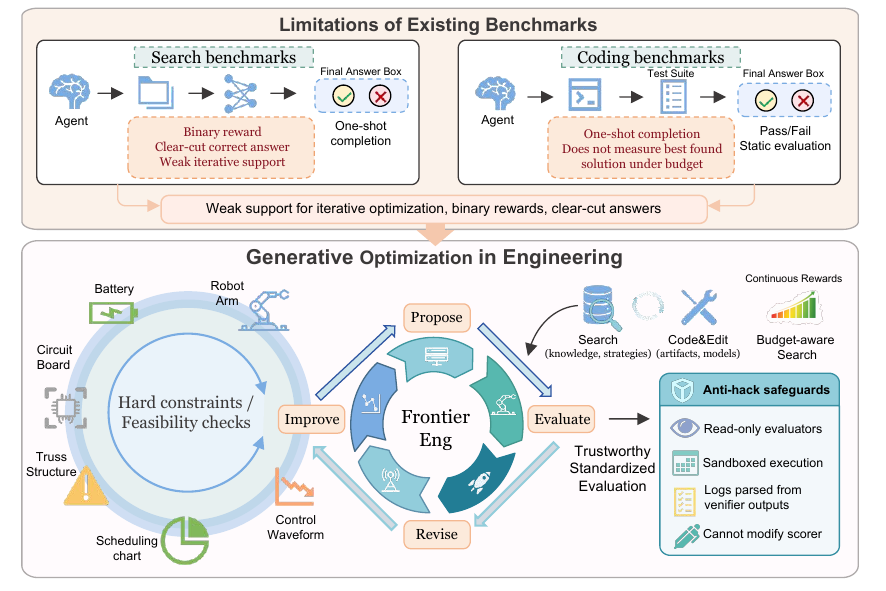}
\caption{Overview of Frontier-Eng. \textbf{Top:} We contrast binary-reward agent benchmarks (search/coding with pass--fail style outcomes) against \emph{generative optimization}, where an agent repeatedly proposes code edits, receives verifier feedback, and improves under a fixed interaction budget. \textbf{Bottom left:} Frontier-Eng covers $47$ tasks across five engineering categories, spanning heterogeneous artifact types, objective structures, and simulator families. \textbf{Bottom right:} The integration pipeline enforces read-only evaluators, isolated execution, feasibility validation, and verifier-parsed scoring, so gains must come from genuine solution improvement rather than reward hacking.}
\label{fig:overview}
\end{figure*}

The path from concept to impact in engineering follows a well-known iterative cycle: define requirements, build a prototype, test against constraints, and refine \citep{Grote2009SpringerHO,Blockley2012EngineeringAV}. While the initial creation of a working solution receives the most attention, experienced practitioners recognize that \emph{optimization}---the systematic, iterative improvement of a feasible design under real-world constraints---is where most of the value is ultimately captured. A charging algorithm that reduces battery charging time by ten minutes, a truss topology that saves fifteen percent of structural material while meeting safety codes, a scheduling heuristic that cuts factory makespan by twenty percent: these incremental improvements compound into enormous practical and economic value. Optimization is not a finishing touch; it is the core of engineering.

Yet today's most capable AI agents and their corresponding benchmarks remain largely oriented toward \emph{0-to-1} tasks with binary outcomes. Search agents retrieve definitive answers to factual or analytical questions by traversing knowledge sources on the web \citep{Wei2025BrowseCompAS,Phan2025HumanitysLE,mialon2023gaia}. Coding agents generate programs that either pass or fail a test suite \citep{Jimenez2023SWEbenchCL,chowdhury2024swebenchverified,Jain2024LiveCodeBenchHA}. These settings share a common structure: there exists a clear-cut correct answer, and the reward is essentially binary---pass or fail. However, a vast and practically consequential class of problems operates in a fundamentally different regime. In real-world engineering, the goal is rarely to produce a single correct artifact from scratch; rather, an initial feasible solution already exists, and the challenge is to \emph{iteratively optimize} it under domain-specific constraints. This process resembles research itself: it requires \emph{searching}---retrieving relevant domain skills and identifying promising strategies---and \emph{coding}---translating those strategies into executable implementations that can be evaluated by simulators, solvers, or rule-based verifiers. Neither skill alone is sufficient; effective optimization demands their tight integration within a closed feedback loop. Moreover, many engineering optimization problems have no known theoretical optimum---a GPU kernel can always be made faster, a scheduling heuristic can always be made tighter, a control policy can always be made more robust---so the search space is effectively unbounded and continued effort continues to yield measurable gains. Even for problems where an optimum exists in principle, it is often unreachable in practice, making the operative question not ``is the solution correct?'' but ``how good a solution can the agent find within budget?''

At the same time, recent work suggests that large language models are beginning to exhibit an emerging capacity for iterative optimization and discovery. FunSearch demonstrated that LLM-guided program search can yield novel mathematical constructions \citep{DBLP:journals/nature/RomeraParedesBNBKDREWFKF24}. AlphaEvolve extended this paradigm to broader scientific and algorithmic discovery via evolutionary code editing under automated evaluation \citep{novikov2025alphaevolve}. Learning to Discover at Test Time further showed that models can continue adapting at test time to a fixed discovery problem, achieving new state-of-the-art solutions across mathematics, GPU kernel engineering, algorithm design, and biology \citep{yuksekgonul2026learning}. Self-Refine formalized iterative self-feedback as a general propose--critique--revise paradigm \citep{Madaan2023SelfRefineIR}. More broadly, recent work has articulated the possibility of automating larger portions of the research loop with coordinated LLM agents \citep{liu2025vision}. We refer to this family of methods collectively as \emph{generative optimization}: the use of generative models as iterative proposers within an evaluate-and-improve loop. Unlike conventional machine learning benchmarks that separate training data from held-out evaluation, generative optimization challenges a model to \emph{self-evolve} \citep{gao2025survey} on a fixed problem instance, searching for progressively better solutions under a finite interaction budget.

However, existing generative optimization efforts have largely targeted narrow problem domains---mathematical function discovery, algorithm design for specific combinatorial problems, or synthetic optimization landscapes \citep{DBLP:journals/nature/RomeraParedesBNBKDREWFKF24,Liu2024EvolutionOH,Yang2023LargeLM}. These settings primarily test pure reasoning or heuristic generation in isolation, rather than the full pipeline of \emph{domain knowledge retrieval, constrained code synthesis, and iterative refinement under realistic verifier feedback}. Crucially, no comprehensive benchmark exists to evaluate generative optimization agents across the breadth of real engineering disciplines. We argue that modern engineering provides a natural and practically valuable testbed for this capability. Engineering spans electrical, mechanical, aerospace, civil, and computer disciplines and underpins critical infrastructure and much of modern life \citep{Chen2004TheEE,Grote2009SpringerHO,Chen2002TheCE,Blockley2012EngineeringAV}. At its core, engineering design and optimization translate requirements and constraints into executable artifacts---controllers, circuits, structures, and system configurations---whose quality is determined by feasibility and performance under domain-specific evaluation. Every improvement on these tasks carries tangible real-world value: faster computations, safer structures, more efficient processes, and reduced environmental impact.

To illustrate concretely, consider the \emph{battery fast-charging} task included in our benchmark. A lithium-ion cell must be charged from low to target state-of-charge as quickly as possible, but the charging profile---a sequence of current stages and SOC switch points---is constrained by hard safety limits on voltage, temperature, lithium plating, and long-term degradation. The agent starts from a naive constant-current baseline and must discover a multi-stage profile that navigates the tradeoff between charging speed, thermal safety, and battery longevity. Evaluation is performed by a reduced-order electrochemical--thermal--aging simulator whose parameters are fixed and whose source code is read-only; the agent cannot game the score without actually improving the charging physics. This single task already demands electrochemical domain knowledge (what causes plating? how does temperature affect aging?), algorithmic reasoning (how to structure the current stages?), and iterative code refinement (adjust switch points based on simulation feedback)---precisely the combination of searching and coding that generative optimization must master.

To systematically evaluate this capability at scale, we introduce \textbf{Frontier-Eng}, a large-scale benchmark for generative optimization agents on real-world engineering tasks. Frontier-Eng contains $47$ tasks grouped into five engineering categories: computing and quantum information, operations research and decision science, robotics and control, optics and communication systems, and physical sciences and engineering design. Tasks are sourced from established engineering competitions, academic benchmark suites, classic coursework, industrial simulation tools, and original contributions by domain experts. Each task packages an editable artifact, an executable verifier with hard feasibility constraints, a contributor-provided baseline, and lightweight metadata behind a unified agent-facing interface. To prevent reward hacking, all evaluators and reference data are marked read-only: scoring is performed by independent, frozen verifiers---FEM solvers, physics simulators, OpenSSL reference implementations, or black-box emulators---that the agent cannot modify. Candidates are executed in sandboxed temporary directories, and output metrics are parsed from verifier-produced logs rather than self-reported by the agent.

% TODO: Add a paragraph summarizing key experimental findings once the experiments section is finalized.

Frontier-Eng makes three contributions:
\begin{enumerate}
\item We formalize \emph{generative optimization} as a distinct evaluation scope for AI agents---one that requires iterative, budget-aware improvement of executable artifacts under hard engineering constraints, rather than one-shot answer generation or binary pass/fail completion.
\item We introduce a benchmark of $47$ real-world engineering tasks across five broad categories with a unified, metadata-driven evaluation interface that preserves domain-specific simulators and verifiers while supporting standardized cross-task comparison through rank-based aggregation and win rate.
q\item We evaluate representative generative optimization methods---spanning evolutionary search, sample-efficient evolution, and tree-based exploration---across multiple frontier language models, establishing initial scaling trends and identifying systematic failure modes that point to concrete directions for future agent design.
\end{enumerate}

\section{Introducing the Frontier-Eng Benchmark}
% Frontier-Eng is a benchmark suite rather than a static dataset: each task couples an agent-facing problem statement with an executable environment that evaluates candidate artifacts. Frontier-Eng currently contains $47$ tasks across five broad engineering categories. The purpose of the benchmark is to compare budgeted search for high-quality feasible solutions across heterogeneous engineering problems under a shared protocol, rather than to catalog repository internals.

\subsection{Task formulation and generative optimization}
\label{sec:formulation}

\paragraph{Engineering optimization tasks.}
An engineering optimization task can be described as a triple
\[
\tau = (\mathcal{C},\; x_0,\; \mathcal{E}).
\]
The \emph{task context} $\mathcal{C}$ encodes everything an agent may read but not alter: the problem specification, constraint descriptions, reference data, and any supporting code or documentation. The \emph{initial solution} $x_0 \in \mathcal{X}$ is a feasible but potentially naive starting artifact---a piece of source code, a configuration file, or a structured submission---that the agent will iteratively improve. The \emph{evaluator} $\mathcal{E}: \mathcal{X} \to \{0,1\} \times \mathbb{R}$ is a fixed function, external to and unmodifiable by the agent, that takes a candidate solution $x$ and returns two signals:
\begin{itemize}
\item a \emph{feasibility indicator} $v(x) \in \{0,1\}$, reflecting whether $x$ satisfies all hard constraints (safety limits, structural integrity, correctness checks, etc.);
\item a \emph{scalar score} $s(x) \in \mathbb{R}$ (higher is better), meaningful only when $v(x)=1$.
\end{itemize}
This formulation is intentionally general. The solution space $\mathcal{X}$ may consist of Python scripts, C source files, CUDA kernels, or structured parameter submissions; the evaluator $\mathcal{E}$ may be a physics simulator, a finite-element solver, a cryptographic reference check, or a black-box emulator. The abstraction imposes no assumptions on the internal structure of either---it requires only that the agent can submit candidates and receive grounded feedback. When the evaluator involves internal randomness (e.g., Monte Carlo simulation), $s(x)$ denotes the expected score, approximated in practice by averaging over repeated runs with fixed seeds.

\paragraph{Generative optimization.}
Given a task $\tau$ and an interaction budget $B$, a \emph{generative optimization} agent $\mathcal{A}$ solves the following iterative problem. Let $(v_0, s_0) = \mathcal{E}(x_0)$ be the evaluation of the initial solution. At each step $t = 1, \ldots, B$:
\begin{enumerate}
\item The agent proposes a new candidate conditioned on the task context and the full history of prior attempts:
\[
x_t = \mathcal{A}\!\left(\mathcal{C},\; H_{t-1}\right), \quad \text{where } H_{t-1} = \bigl\{(x_i, v_i, s_i)\bigr\}_{i=0}^{t-1}.
\]
\item The evaluator returns $(v_t, s_t) = \mathcal{E}(x_t)$. Both feasible and infeasible steps consume budget.
\end{enumerate}
The objective is to maximize the best feasible score found within budget:
\[
s^{\star} = \max_{\,0 \le t \le B,\; v_t = 1}\; s_t.
\]

What distinguishes generative optimization from classical mathematical optimization is the nature of the agent $\mathcal{A}$. Rather than performing gradient descent or sampling in a continuous parameter space, a generative optimization agent is built around a large language model that produces each candidate $x_t$ through \emph{code generation}: it reads the task context and prior evaluation feedback as natural-language and structured input, reasons about what to change, and emits a revised artifact. The full history $H_{t-1}$ is available in principle; in practice, different search strategies---evolutionary selection, tree-based exploration, bandit-guided sampling---determine which subset of $H_{t-1}$ is surfaced to the LLM as prompt context at each step. This separation between the \emph{search strategy} (how to select and present history) and the \emph{proposal mechanism} (the LLM that generates code) is a defining architectural feature of current generative optimization systems.

\subsection{Benchmark construction and integrity}
\label{sec:construction}

Frontier-Eng instantiates the formulation above across $47$ engineering tasks spanning diverse subfields. This section describes how tasks are sourced, what inclusion criteria they must satisfy, and what safeguards prevent reward hacking.

\paragraph{Task sources.}
Tasks are drawn from five complementary channels to ensure both breadth and quality: (1)~\emph{established engineering competitions} such as the ISCSO structural optimization challenges, whose problem data, constraint definitions, and scoring rubrics are publicly archived; (2)~\emph{academic benchmark suites} including Summit reaction emulators, MQT Bench quantum circuits, and the OpenProblems single-cell analysis platform; (3)~\emph{classic coursework} such as the CS:APP \texttt{MallocLab}, which provides well-understood invariants and a mature test harness; (4)~\emph{industrial-grade simulation tools} including MuJoCo, PyBullet, and the \texttt{SustainDC} data-center environment; and (5)~\emph{original contributions} by domain experts, covering areas such as adaptive optics, fiber-network optimization, inventory management, and battery electrochemistry. Each task is accompanied by a contributor-written problem statement, a runnable evaluator, and a feasible initial solution.

\paragraph{Inclusion criteria.}
A candidate task is admitted into Frontier-Eng only if it satisfies four requirements: (i)~it has a clear, self-contained specification that an agent can read without external resources; (ii)~it exposes an editable artifact $x_0$ in a well-defined solution space; (iii)~it provides a runnable evaluator $\mathcal{E}$ that returns both a feasibility indicator and a scalar score under a declared runtime environment (conda, Docker, or system Python); and (iv)~the initial solution $x_0$ is verified to be feasible, i.e., $v(x_0) = 1$, so that every agent begins from a valid starting point.

\paragraph{Quality control.}
Task contributions follow a two-stage review process. First, an automated agent review checks code standards, evaluator executability, and basic interface compliance. Second, human maintainers verify the engineering soundness of the problem formulation, the correctness of the evaluator, and the adequacy of the constraint definitions. Tasks that fail either stage are returned for revision before inclusion.

\paragraph{Evaluation integrity.}
To prevent reward hacking---agents exploiting evaluator weaknesses rather than genuinely improving solutions---Frontier-Eng enforces three layers of safeguards:
\begin{itemize}
\item \textbf{Isolation.} All evaluator code, reference data, and verification scripts are marked read-only and cannot be modified by the agent. Candidate solutions execute in sandboxed temporary directories that contain only the files explicitly declared as agent-accessible.
\item \textbf{Verifier-parsed scoring.} Scores are extracted from the evaluator's own output logs---stdout, structured JSON, or simulator-produced artifacts---rather than from any file the candidate writes. The agent cannot self-report a high score; it must earn one from the frozen verifier.
\item \textbf{Evaluation robustness.} Where applicable, evaluators employ multiple test seeds, randomized inputs, or multi-scenario averaging to reduce the risk of shortcut exploitation. Correctness checks (e.g., OpenSSL reference comparison, schedule validation, FEM re-solve) ensure that claimed improvements correspond to genuine physical or algorithmic gains.
\end{itemize}

%% ─────────────────────────────────────────────────────────────────────
\subsection{Benchmark composition and coverage}
\label{sec:composition}

Frontier-Eng contains $47$ tasks drawn from diverse engineering subfields. To organize the presentation and facilitate cross-domain analysis, we group these tasks into five broad \emph{engineering categories} (Table~\ref{tab:benchmark_overview}).

\begin{table*}[!ht]
\caption{Complete task inventory of Frontier-Eng. The $47$ tasks are grouped into five engineering categories. A detailed per-task catalog including scoring formulas and anti-hack measures is provided in Appendix~\ref{app:task_catalog}.}
\label{tab:benchmark_overview}
\centering
\small
\begin{adjustbox}{max width=.95\textwidth}
\begin{tabular}{@{}>{\texttt}l >{\texttt}l p{9.2cm}@{}}
\toprule
\multicolumn{1}{@{}l}{Subfield} & \multicolumn{1}{l}{Task} & Description \\
\midrule
\multicolumn{3}{@{}l}{\textbf{Computing \& Quantum Information (10 tasks)}} \\
\addlinespace[2pt]
KernelEngineering & FlashAttention & Optimize causal scaled dot-product attention CUDA kernel \\
KernelEngineering & MLA & Optimize multi-head latent attention CUDA kernel \\
KernelEngineering & TriMul & Optimize triangular multiplicative update CUDA kernel \\
ComputerSystems & MallocLab & High-performance C memory allocator (utilization/throughput) \\
Cryptographic & AES-128 & C++ AES-128 CTR throughput (OpenSSL verified) \\
Cryptographic & SHA-256 & C++ SHA-256 throughput (OpenSSL verified) \\
Cryptographic & SHA3-256 & C++ SHA3-256 throughput (OpenSSL verified) \\
QuantumComputing & Routing QFTEntangled & Optimize QFT circuit routing on IBM Falcon (gate/depth) \\
QuantumComputing & Clifford+T Synthesis & Clifford+T synthesis for QFT (T-gate/depth) \\
QuantumComputing & Cross-Target QAOA & Joint QAOA optimization for IBM and IonQ backends \\
\midrule
\multicolumn{3}{@{}l}{\textbf{Operations Research \& Decision Science (9 tasks)}} \\
\addlinespace[2pt]
InventoryOpt & tree\_gsm\_safety\_stock & Optimize safety stock on tree-structured supply network \\
InventoryOpt & general\_meio & Base-stock optimization for multi-echelon network \\
InventoryOpt & joint\_replenishment & Cycle time and order multiples for 8 SKUs \\
InventoryOpt & finite\_horizon\_dp & Time-varying (s,S) policy for 8-period inventory \\
InventoryOpt & disruption\_eoqd & Order quantity optimization under supply disruptions \\
JobShop & abz & Minimize JSSP makespan (ABZ family, up to 20$\times$15) \\
JobShop & swv & Minimize JSSP makespan (SWV family, up to 50$\times$10) \\
JobShop & ta & Minimize JSSP makespan (Taillard family, up to 100$\times$20) \\
PyPortfolioOpt & robust\_mvo\_rebalance & Robust MVO rebalancing with sector/factor constraints \\
\midrule
\multicolumn{3}{@{}l}{\textbf{Robotics, Control \& Energy Systems (8 tasks)}} \\
\addlinespace[2pt]
Robotics & DynObstacleNav & Robot path planning avoiding dynamic obstacles \\
Robotics & PIDTuning & Tune 12 PID gains for a 2D quadrotor \\
Robotics & QuadrupedGait & Optimize MuJoCo Ant gait parameters for speed \\
Robotics & RobotArmCycleTime & Minimize KUKA arm motion time with collision avoidance \\
Robotics & UAVInspection & UAV inspection coverage under wind and no-fly zones \\
EnergyStorage & BatteryFastChargingProfile & Multi-stage CC charging under thermal/plating limits \\
EnergyStorage & BatteryFastChargingSPMe & Charging optimization with high-fidelity SPMe model \\
SustainableDC & hand\_written\_control & Joint load-shifting, cooling, and battery control \\
\midrule
\multicolumn{3}{@{}l}{\textbf{Optics \& Communication Systems (10 tasks)}} \\
\addlinespace[2pt]
Optics & adaptive\_fault\_tolerant\_fusion & Wavefront sensor slope fusion for AO control \\
Optics & adaptive\_temporal\_smooth & Sequential AO control with command smoothness \\
Optics & phase\_dammann\_uniform & Binary phase optimization for Dammann grating \\
Optics & phase\_fourier\_holography & 2D phase pattern for Fourier hologram \\
Optics & fiber\_wdm\_channel\_power & WDM channel and power allocation for 14 users \\
Optics & fiber\_mcs\_power\_scheduling & Joint MCS and power allocation for 22 users \\
Optics & fiber\_guardband\_packing & Spectrum packing with guard-bands and BER constraints \\
Optics & holographic\_multifocus\_ratio & Phase design for target multi-focus power ratios \\
Optics & holographic\_multiplane & Multi-plane holographic focusing (efficiency/ratio) \\
WirelessChannel & HighReliableSimulation & Importance-sampling BER estimator for Hamming codes \\
\midrule
\multicolumn{3}{@{}l}{\textbf{Physical Sciences \& Engineering Design (10 tasks)}} \\
\addlinespace[2pt]
StructuralOpt & ISCSO2015 & Minimize weight of 2D truss under stress/displacement \\
StructuralOpt & ISCSO2023 & Minimize weight of 3D tower with discrete sections \\
StructuralOpt & TopologyOptimization & Minimize compliance of 2D MBB beam (volume constraint) \\
ReactionOpt & snar\_multiobjective & Pareto-optimize SnAr reaction (yield vs. environment) \\
ReactionOpt & mit\_case1\_mixed & Maximize reaction yield with mixed variables \\
ReactionOpt & reizman\_suzuki\_pareto & Pareto-optimize Suzuki coupling (yield vs. turnover) \\
Astrodynamics & MannedLunarLanding & Maximize CRTBP lunar payload (Octave validation) \\
Aerodynamics & CarAerodynamicsSensing & 30 sensor locations for pressure field reconstruction \\
SingleCellAnalysis & predict\_modality & Cross-modality gene expression prediction (RNA to ADT) \\
EngDesign & EngDesign (7 sub-problems) & Multi-task: drivers, denoising, CPU logic, path planning \\
\bottomrule
\end{tabular}
\end{adjustbox}
\end{table*}

\paragraph{Computing \& Quantum Information (10 tasks).}
Tasks in this category require optimizing executable code or circuit-level representations for throughput, latency, or structural cost. GPU kernel engineering tasks (\texttt{FlashAttention}, \texttt{MLA}, \texttt{TriMul}) ask the agent to write CUDA or Triton kernels whose correctness is verified against a reference implementation and whose performance is measured by wall-clock latency. \texttt{MallocLab} evaluates a C dynamic memory allocator on utilization and throughput across allocation traces. Three cryptographic tasks (\texttt{AES-128}, \texttt{SHA-256}, \texttt{SHA3-256}) measure C++ implementation throughput after OpenSSL-backed correctness checks. Three quantum computing tasks optimize circuit routing, Clifford+T synthesis, and cross-target QAOA compilation, scored by gate count and depth after canonical transpilation against MQT Bench references.

\paragraph{Operations Research \& Decision Science (9 tasks).}
These tasks involve discrete or combinatorial optimization problems with well-defined cost models. Five inventory optimization tasks span tree-structured safety stock, general multi-echelon simulation, joint replenishment, finite-horizon dynamic programming, and disruption-aware EOQ, each scored by weighted composites of cost, service level, and robustness. Three job-shop scheduling families (ABZ, SWV, TA, covering instances up to $100 \times 20$) score the ratio of achieved makespan to known optima or upper bounds; solutions are constrained to pure Python without external solvers. A robust portfolio rebalancing task evaluates mean-variance optimization under sector, factor, and turnover constraints against a CVXPY reference.

\paragraph{Robotics, Control \& Energy Systems (8 tasks).}
The agent designs controllers, planners, or operational policies for dynamical systems. Robotics tasks include differential-drive navigation with dynamic obstacles, cascaded PID tuning for a quadrotor, quadruped gait parameter optimization in MuJoCo, KUKA arm cycle-time minimization with collision avoidance in PyBullet, and UAV inspection coverage under wind disturbances. Two battery fast-charging tasks optimize multi-stage current profiles under electrochemical, thermal, and degradation constraints at different model fidelities. A sustainable data-center task requires a joint load-shifting, cooling, and battery-dispatch policy evaluated against a noop reference over fixed scenarios.

\paragraph{Optics \& Communication Systems (10 tasks).}
Nine optics tasks span four sub-areas: adaptive optics control (fault-tolerant multi-WFS fusion; temporally smooth control under delay and plant mismatch), diffractive optical element design (Dammann grating uniformity; Fourier pattern holography), fiber-network optimization (WDM channel allocation; MCS-power scheduling; guard-band spectrum packing), and holographic focusing (multi-focus power ratio; multi-plane focusing). Each task is scored by physics-based metrics---Strehl ratio, BER, diffraction efficiency, or pattern fidelity---computed from wave-optics or link-budget simulators. A wireless channel simulation task evaluates importance-sampling BER estimators for Hamming codes, scored on both accuracy and runtime.

\paragraph{Physical Sciences \& Engineering Design (10 tasks).}
This category collects tasks grounded in physical simulation or multi-domain design tools. Three structural optimization tasks (\texttt{ISCSO2015} 2D truss, \texttt{ISCSO2023} 3D tower, \texttt{TopologyOptimization} MBB beam) minimize weight or compliance under stress and displacement constraints verified by built-in FEM solvers. Three reaction optimization tasks use Summit emulators for \texttt{snar\_multiobjective}, \texttt{mit\_case1\_mixed}, and \texttt{reizman\_suzuki\_pareto}, scored by hypervolume or best yield. \texttt{MannedLunarLanding} maximizes payload under CRTBP dynamics validated by Octave integration. \texttt{CarAerodynamicsSensing} optimizes sensor placement for pressure-field reconstruction using a frozen neural surrogate. \texttt{predict\_modality} predicts cross-modality gene expression, scored by correlation and RMSE against held-out data. Finally, \texttt{EngDesign} bundles seven heterogeneous sub-problems (device drivers, image denoising, CPU control logic, robot path planning, and topology optimization) into a single multi-task submission.

\paragraph{Diversity dimensions.}
Beyond disciplinary breadth, Frontier-Eng exhibits diversity along several axes relevant to agentic optimization. \emph{Artifact languages} include Python (majority), C, C++, and CUDA/Triton. \emph{Verifier types} range from physics simulators and FEM solvers, through black-box emulators and cryptographic reference checks, to MuJoCo/PyBullet rollouts. \emph{Optimization structures} include single-objective minimization, multi-objective Pareto search, constrained feasibility problems, and combinatorial scheduling. \emph{Compute requirements} span CPU-only tasks, GPU-mandatory kernel and robotics tasks, and Docker-isolated engineering design evaluation. This heterogeneity ensures that no single search strategy or prompting approach is universally advantageous, supporting meaningful comparison of agentic methods.

\begin{figure}[t]
\centering
\includegraphics[width=0.94\linewidth]{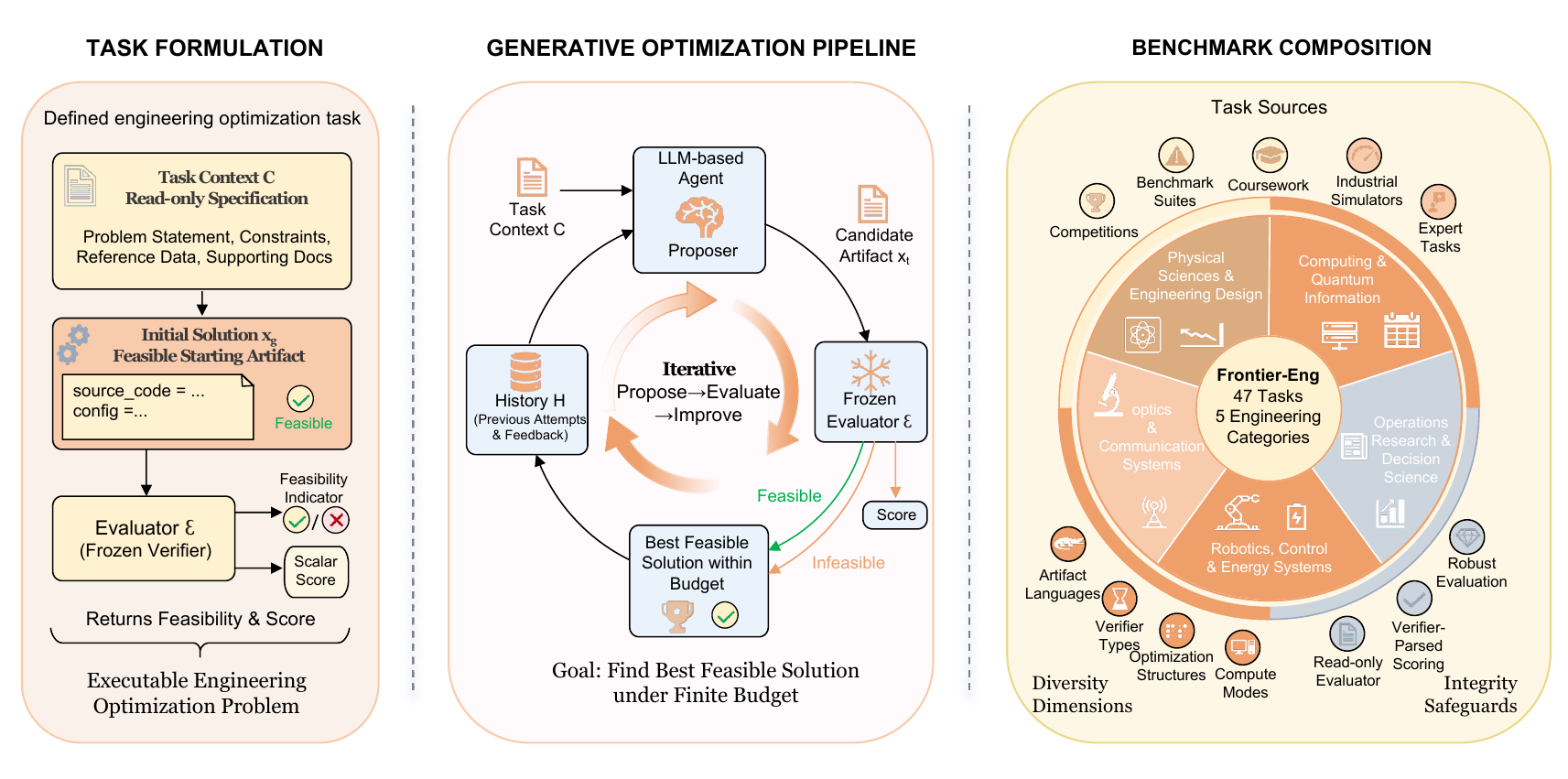}
\caption{Method and benchmark composition overview of Frontier-Eng. The figure summarizes the unified task interface $\tau=(\mathcal{C},x_0,\mathcal{E})$, the iterative propose--evaluate--improve loop under a fixed budget, and the benchmark-level aggregation pipeline for cross-task comparison; it also shows the data composition of the benchmark, which contains 47 tasks organized into five engineering categories.}
\label{fig:benchmark_composition}
\end{figure}

\subsection{Evaluation protocol}
\label{sec:eval_protocol}

Comparing generative optimization agents across heterogeneous engineering tasks poses a fundamental challenge: raw scores are measured in incompatible units---throughput in Mbps for cryptographic tasks, makespan in time units for scheduling, structural weight in kilograms, Strehl ratio for optics---and their scales, ranges, and baseline difficulties vary by orders of magnitude. No single normalization can map all tasks to a common absolute scale without strong assumptions. Frontier-Eng therefore adopts a multi-level evaluation protocol that combines a rank-based primary metric with a distributional analysis and supplementary diagnostics.

\paragraph{Primary metric: Average Rank.}
For each task $i$ and method $m$, let $s_{i,m}^{\star}$ denote the best feasible score obtained within budget $B$. Since every initial solution $x_0$ is guaranteed feasible, every method achieves at least $s_{i,m}^{\star} \ge s_{i,0}$, the score of the starting point. We rank all $M$ methods on each task by their best feasible score (higher is better), assigning rank $1$ to the best and rank $M$ to the worst, with ties receiving averaged ranks. The \emph{average rank} of method $m$ across all $N$ tasks is simply
\[
R_m = \frac{1}{N} \sum_{i=1}^{N} \mathrm{rank}_{i,m}, \quad \mathrm{rank}_{i,m} \in \{1, 2, \dots, M\}.
\]
Lower is better: $R_m = 1$ means that method $m$ ranks first on every task. This metric is unit-free, treats all tasks equally, and does not require knowledge of theoretical optima or reference scores. Because it aggregates ordinal positions directly, it also provides an intuitive ``pseudo-rank'' interpretation: $R_m = 2.3$ means that the method places, on average, between second and third across the benchmark. Its limitation is that it discards magnitude: a method that wins by a negligible margin and one that wins by a large margin receive the same rank credit.

\paragraph{Distributional analysis: Performance Profile.}
To recover the magnitude information that ranks discard, we adopt the performance profile framework of \citet{Dolan2002BenchmarkingOA}. Because Frontier-Eng contains tasks with negative scores (e.g., structural weight minimization, where scores are reported as negative weights), the standard ratio $s_{i,\text{best}}^{\star}/s_{i,m}^{\star}$ is undefined or sign-inverted for non-positive values. We therefore apply a per-task translation before computing ratios. For each task $i$, let $s_{i,\min}^{\star} = \min_m s_{i,m}^{\star}$ be the worst score achieved by any method on that task. Define the translated score
\[
\tilde{s}_{i,m} = s_{i,m}^{\star} - s_{i,\min}^{\star} + 1,
\]
which shifts all scores to be strictly positive while preserving within-task ordering. The performance ratio is then
\[
\rho_{i,m} = \frac{\tilde{s}_{i,\text{best}}}{\tilde{s}_{i,m}}, \quad \tilde{s}_{i,\text{best}} = \max_m \tilde{s}_{i,m}.
\]
By construction $\rho_{i,m} \ge 1$, with equality when method $m$ achieves the best translated score. The performance profile of method $m$ is the cumulative distribution
\[
P_m(\alpha) = \frac{1}{N}\,\bigl|\bigl\{i : \rho_{i,m} \le \alpha\bigr\}\bigr|, \quad \alpha \ge 1.
\]
$P_m(1)$ is the fraction of tasks on which method $m$ achieves the best score (invariant to the translation). For $\alpha > 1$, the interpretation is relative to the translated scale: $\alpha = 1.1$ means the translated score is within $1/1.1 \approx 90.9\%$ of the translated best, rather than exactly $90.9\%$ of the original best score. This semantic shift is a necessary cost of supporting mixed-sign benchmarks; the within-task translation ensures that the worst observed performance sets the denominator anchor rather than an arbitrary zero. A method whose profile rises steeply and stays high is both competitive and consistent; one with a high $P_m(1)$ but slow rise is strong on some tasks but unreliable on others.

\paragraph{Supplementary metric: Win Rate.}
As an additional diagnostic, we report the \emph{win rate over baseline}: the fraction of tasks on which a method's best feasible score strictly exceeds the initial solution's score, $s_{i,m}^{\star} > s_{i,0}$. This captures the reliability of improvement---how often the agent manages to improve upon the starting point at all---without measuring by how much. A high average rank with a low win rate would indicate that a method achieves large gains on a few tasks but fails to improve on many others.

\paragraph{Category-level breakdown.}
Because the five engineering categories (Table~\ref{tab:benchmark_overview}) group tasks with related skill requirements, we also report per-category versions of the above metrics. This breakdown reveals whether a method's aggregate strength comes from uniform competence or from dominance in a single category, and identifies category-specific failure modes that aggregate numbers would obscure.

\paragraph{Protocol separation.}
The evaluation protocol is deliberately independent of any particular search algorithm. The same task interface, verifier semantics, and interaction budget apply whether the agent uses evolutionary search, tree-based exploration, simple iterative refinement, or any future strategy. This separation ensures that Frontier-Eng remains a stable benchmark as generative optimization methods continue to evolve.

\subsection{Case study: \textsc{SustainableDataCenterControl}}
\label{sec:case_study}

To ground the formulation in a concrete instance, we walk through the \textsc{SustainableDataCenterControl} task \citep{Naug2024SustainDCB}, which exemplifies the multi-objective, constraint-rich structure typical of engineering optimization.

\begin{figure}[t]
\centering
\includegraphics[width=\linewidth]{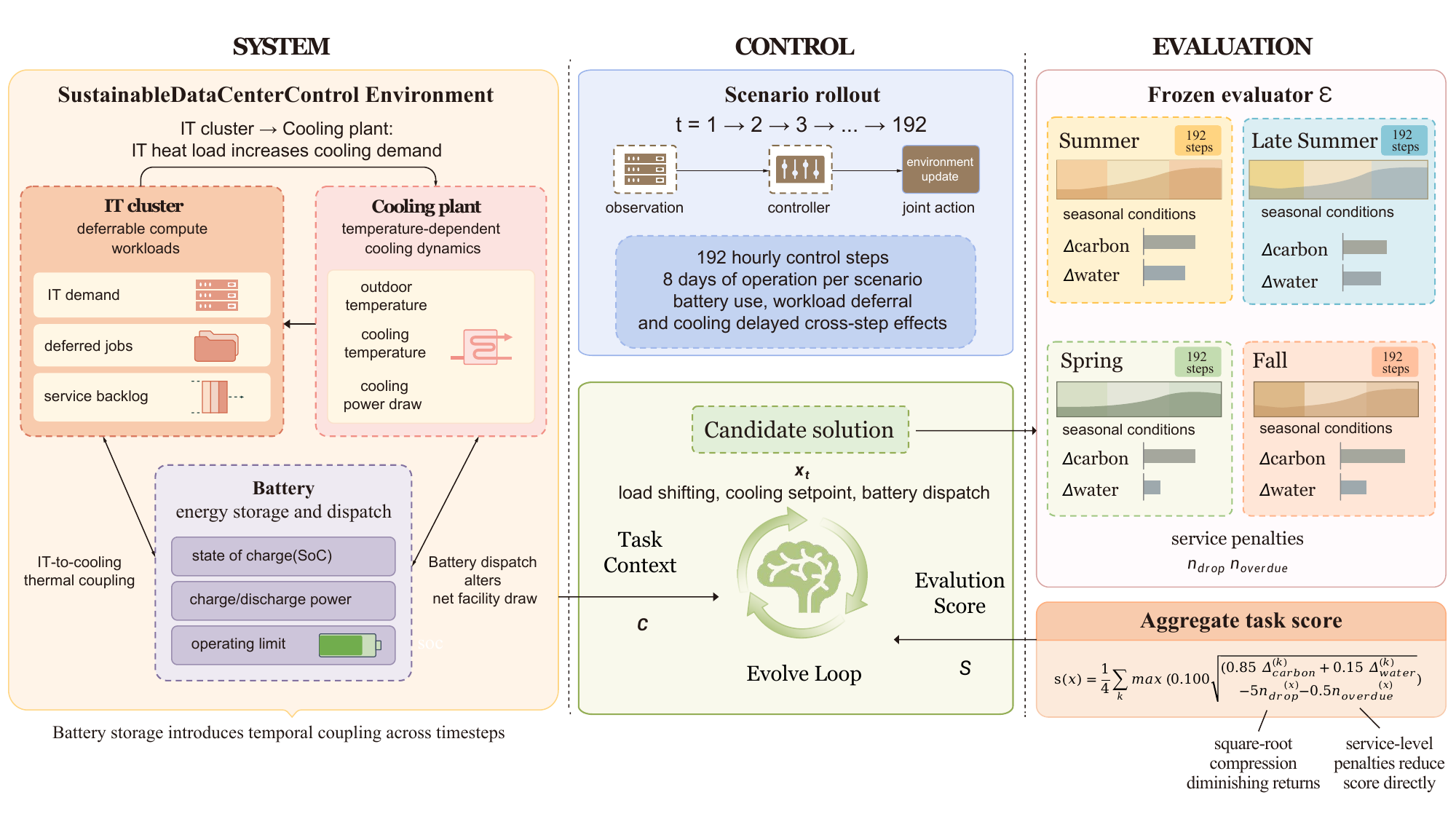}
\caption{Case study under \texttt{openevolve} for \textsc{SustainableDataCenterControl}. The figure illustrates how policies evolve from the contributor baseline through repeated propose--evaluate--improve iterations under a fixed verifier and budget. It highlights representative optimization trajectories and final outcomes across model families, emphasizing three practical signals: improvement speed in early iterations, achievable feasible gain over baseline, and stability of late-stage refinements.}
\label{fig:sustaindc_example}
\end{figure}

\paragraph{Task formulation.} Following the triple $\tau = (\mathcal{C},\, x_0,\, \mathcal{E})$ introduced in Section~\ref{sec:formulation}:

\begin{itemize}
\item \textbf{Task context $\mathcal{C}$.} The data center operates three tightly coupled subsystems: an \emph{IT cluster} that can defer or redistribute computational workloads across time steps (load shifting), a \emph{cooling plant} whose power draw depends on outdoor temperature and internal heat load, and a \emph{battery} that can store cheap or renewable energy for later use. At each time step, the agent observes a partial state vector comprising current IT demand, outdoor temperature, grid carbon intensity, battery state-of-charge, and cooling-system temperatures. The context $\mathcal{C}$ exposes the full simulator dynamics, the observation and action-space definitions, and four fixed evaluation scenarios representing distinct seasonal and carbon-intensity profiles---for example, a summer week with high cooling load and volatile renewable supply versus a winter week with low cooling demand but persistently high grid carbon.

\item \textbf{Initial solution $x_0$.} The starting policy is a stateless no-op controller: it defers no workloads, applies only the minimum cooling required to prevent thermal shutdown, and holds the battery at its initial charge level. This policy is guaranteed feasible---it never violates action bounds or causes runtime failures---but it captures none of the potential savings from coordinated scheduling, proactive cooling, or strategic battery cycling.

\item \textbf{Evaluator $\mathcal{E}$.} For each of the four scenarios, the evaluator executes a closed-loop rollout of $192$ time steps (each representing one hour of real-world operation, totaling eight days per scenario). At every step, the candidate policy maps the current observation to a joint action across the three subsystems; the simulator advances the physical state and accumulates carbon emissions (kg) and water consumption (liters). After all four rollouts, the evaluator reruns the same scenarios with the no-op reference and computes per-scenario improvement fractions. The feasibility flag $v(x)$ is $1$ only if no action-space violation or unhandled exception occurs in any scenario. The scalar score is
\[
s(x) = \frac{1}{4}\sum_{k=1}^{4} \max\!\Big(0,\; 100\sqrt{0.85\,\Delta_{\text{carbon}}^{(k)} + 0.15\,\Delta_{\text{water}}^{(k)}} \;-\; 5\,n_{\text{drop}}^{(k)} \;-\; 0.5\,n_{\text{overdue}}^{(k)}\Big),
\]
where $\Delta_{\text{carbon}}^{(k)}$ and $\Delta_{\text{water}}^{(k)}$ are the fractional reductions relative to the no-op reference in scenario $k$, and $n_{\text{drop}}^{(k)}$, $n_{\text{overdue}}^{(k)}$ count dropped and overdue IT tasks respectively. The square root compresses large improvements to discourage overly aggressive strategies that sacrifice service quality, while the linear penalties ensure that any service-level degradation is directly reflected in the score.
\end{itemize}

\paragraph{Why this task is challenging.}
Effective optimization requires reasoning about temporal coupling (energy stored in the battery now enables lower-carbon dispatch later), multi-objective tradeoffs (aggressive load shifting reduces carbon but risks overdue tasks), and scenario robustness (a policy tuned for summer cooling loads may fail under winter carbon profiles). Across the four scenarios, a candidate policy is tested against $768$ hours of diverse operational conditions. The no-op baseline is easy to beat marginally---any reasonable cooling or battery heuristic yields some carbon savings---but achieving a high score demands a coordinated strategy that jointly optimizes all three subsystems while maintaining service-level guarantees across every scenario. Figure~\ref{fig:sustaindc_example} summarizes the task structure and evaluation loop.

% ══════════════════════════════════════════════════════════════════════
% Original §2.4 (preserved for reference)
% ══════════════════════════════════════════════════════════════════════
% Raw task scores are often not directly comparable across domains, since they may correspond to different physical quantities, scales, and baseline strengths. Frontier-Eng therefore standardizes score direction only and uses within-task comparisons for cross-task aggregation. Let $s_{i,m}^{\star}$ denote the best feasible score obtained by method $m$ on task $i$. Our primary aggregate metric is a normalized per-task rank,
% r_{i,m} = 1 - (rank_{i,m}-1)/(M_i-1)
% where feasible runs are ranked ahead of infeasible runs and ties receive averaged ranks.
% We also report win rate over the contributor baseline and median clipped baseline-relative improvement as secondary summaries.

% TODO: Draft the experiments section after finalizing evaluation settings and results.

\section{Experiments}

\subsection{Different Foundation Models under openevolve}
\label{sec:openevolve-model-comparison}

\subsubsection{Setup and Metrics}

We compare nine foundation models under a fixed \texttt{openevolve} budget of 100 iterations on the Experiment~1 runs. All models operate under the same search framework, start from the same contributor-provided initial programs, and are evaluated by the same frozen task verifiers in the declared task environments. For \texttt{gpt-5.4}, we replace the original run with the full 47-task retest under the same protocol, and recompute all rank-based and distributional summaries from the updated score table.

For evaluation, we follow the benchmark protocol in Section~\ref{sec:eval_protocol}. Specifically, we use \emph{average rank} as the primary cross-task metric, and report \emph{performance profile} and \emph{win rate over baseline} as aggregate diagnostics.

\subsubsection{Results and Analysis}
Table~\ref{tab:openevolve-models-rank-en} reports the within-task rank of each model (1 = best) on the full 47-task set after updating \texttt{gpt-5.4} with the retest scores. The retest materially changes the ordering: \texttt{gpt-5.4} now attains the best average rank (3.54), narrowly ahead of \texttt{claude-opus-4.6} (3.63), followed by \texttt{glm-5} (4.34) and \texttt{deepseek-v3.2} (4.76). Relative to the original \texttt{gpt-5.4} run (average rank 5.68), this rerun shifts \texttt{gpt-5.4} from the lower tier of the table to the top position. This rank-based view remains the primary cross-task comparison because it is unit-free and robust to heterogeneous task scales.

\begin{table*}[!t]
\centering
\caption{Within-task rank comparison of nine foundation models on the full 47-task set under \texttt{openevolve} (100 iterations, same initial programs, same frozen verifiers), with \texttt{gpt-5.4} replaced by its retest results under the same protocol. Header abbreviations are unified as: Opus$=$\texttt{claude-opus-4.6}, Qwen$=$\texttt{qwen3-coder-next}, Seed$=$\texttt{seed-2.0-pro}, GPT$=$\texttt{gpt-5.4}, OSS$=$\texttt{gpt-oss-120b}, GLM$=$\texttt{glm-5}, DS$=$\texttt{deepseek-v3.2}, Grok$=$\texttt{grok-4.20}, Gemini$=$\texttt{gemini-3.1-pro-preview}. Rank 1 denotes the best model on that task; ties receive average ranks. Bold with a superscript star marks the best (lowest) rank in each row. The \textit{Average} row reports mean within-task rank over all 47 tasks (lower is better).}
\label{tab:openevolve-models-rank-en}
\scriptsize
\setlength{\tabcolsep}{4pt}
\renewcommand{\arraystretch}{1.08}
\begingroup
\let\origtextbf\textbf
\renewcommand{\textbf}[1]{\textsuperscript{\scriptsize$\star$}\origtextbf{#1}}
\begin{tabular}{lrrrrrrrrr}
\toprule
Task & Opus & Qwen & Seed & GPT & OSS & GLM & DS & Grok & Gemini \\
\midrule
Aerodynamics / CarAerodynamicsSensing & 8 & \textbf{2} & 8 & 4 & 6 & 5 & \textbf{2} & 8 & \textbf{2} \\
Astrodynamics / MannedLunarLanding & 4 & 8.5 & 6 & 2 & 5 & \textbf{1} & 3 & 8.5 & 7 \\
ComputerSystems / MallocLab & \textbf{1} & 8 & 7 & 9 & 4.5 & 2 & 4.5 & 3 & 6 \\
Cryptographic / AES-128 & 4 & 9 & 8 & \textbf{1} & 3 & 7 & 2 & 5 & 6 \\
Cryptographic / SHA-256 & 4 & 8 & 5 & \textbf{1} & 2 & 6 & 9 & 3 & 7 \\
Cryptographic / SHA3-256 & 5 & 7 & 3 & \textbf{1} & 2 & 4 & 6 & 9 & 8 \\
EnergyStorage / BatteryFastChargingProfile & 2 & 9 & 5 & \textbf{1} & 6 & 3 & 7 & 8 & 4 \\
EnergyStorage / BatteryFastChargingSPMe & 9 & 5 & 8 & \textbf{1} & 4 & 6 & 3 & 7 & 2 \\
EngDesign & 9 & 4.5 & \textbf{2} & 8 & 6 & 4.5 & 7 & \textbf{2} & \textbf{2} \\
InventoryOptimization / disruption\_\allowbreak eoqd & 2 & 9 & 7 & \textbf{1} & 5 & 8 & 4 & 6 & 3 \\
InventoryOptimization / finite\_\allowbreak horizon\_\allowbreak dp & 2 & 9 & 8 & \textbf{1} & 7 & 5 & 4 & 3 & 6 \\
InventoryOptimization / general\_\allowbreak meio & 2 & 8 & 9 & \textbf{1} & 5 & 7 & 3 & 6 & 4 \\
InventoryOptimization / joint\_\allowbreak replenishment & 5.5 & 9 & 5.5 & \textbf{1.5} & \textbf{1.5} & 5.5 & 5.5 & 5.5 & 5.5 \\
InventoryOptimization / tree\_\allowbreak gsm\_\allowbreak safety\_\allowbreak stock & 2 & 6 & 6 & \textbf{1} & 6 & 6 & 6 & 6 & 6 \\
JobShop / abz & \textbf{1} & 9 & 7 & 2 & 8 & 3 & 4 & 5 & 6 \\
JobShop / swv & \textbf{1} & 6 & 7 & 2 & 4 & 3 & 8 & 5 & 9 \\
JobShop / ta & \textbf{1} & 5 & 9 & 3 & 8 & 2 & 7 & 6 & 4 \\
KernelEngineering / FlashAttention & 5 & 7 & 2 & \textbf{1} & 6 & 8 & 4 & 9 & 3 \\
KernelEngineering / MLA & 3 & 8 & 6 & 2 & 4 & 5 & 9 & 7 & \textbf{1} \\
KernelEngineering / TriMul & \textbf{1} & 8 & 5 & 9 & 6 & 3 & 4 & 2 & 7 \\
Optics / adaptive\_\allowbreak fault\_\allowbreak tolerant\_\allowbreak fusion & 5 & 5 & 5 & 9 & 5 & 5 & \textbf{1} & 5 & 5 \\
Optics / adaptive\_\allowbreak temporal\_\allowbreak smooth\_\allowbreak control & 6 & \textbf{2} & \textbf{2} & 8 & \textbf{2} & 9 & 6 & 4 & 6 \\
Optics / fiber\_\allowbreak guardband\_\allowbreak spectrum\_\allowbreak packing & 2.5 & 8 & 8 & \textbf{1} & 5 & 2.5 & 8 & 5 & 5 \\
Optics / fiber\_\allowbreak mcs\_\allowbreak power\_\allowbreak scheduling & 2 & 9 & 3.5 & \textbf{1} & 7.5 & 3.5 & 5 & 7.5 & 6 \\
Optics / fiber\_\allowbreak wdm\_\allowbreak channel\_\allowbreak power\_\allowbreak allocation & 4 & 5 & 7 & \textbf{1} & 8 & 2 & 3 & 6 & 9 \\
Optics / holographic\_\allowbreak multifocus\_\allowbreak power\_\allowbreak ratio & 4 & 6 & 7 & \textbf{1} & 2 & 5 & 3 & 9 & 8 \\
Optics / holographic\_\allowbreak multiplane\_\allowbreak focusing & 4 & 5 & 6 & \textbf{1} & 3 & 8 & 2 & 7 & 9 \\
Optics / phase\_\allowbreak dammann\_\allowbreak uniform\_\allowbreak orders & 2 & 7 & 9 & \textbf{1} & 5 & 4 & 6 & 8 & 3 \\
Optics / phase\_\allowbreak fourier\_\allowbreak pattern\_\allowbreak holography & 2 & 9 & 8 & \textbf{1} & 7 & 4 & 5 & 6 & 3 \\
PyPortfolioOpt / robust\_\allowbreak mvo\_\allowbreak rebalance & 2.5 & 5 & 7 & \textbf{1} & 2.5 & 8 & 6 & 4 & 9 \\
QuantumComputing / task\_\allowbreak 01\_\allowbreak routing\_\allowbreak qftentangled & 2 & 7 & 5 & \textbf{1} & 8.5 & 3 & 6 & 4 & 8.5 \\
QuantumComputing / task\_\allowbreak 02\_\allowbreak clifford\_\allowbreak t\_\allowbreak synthesis & 8 & 3.5 & 3.5 & 6 & 8 & \textbf{1} & 3.5 & 8 & 3.5 \\
QuantumComputing / task\_\allowbreak 03\_\allowbreak cross\_\allowbreak target\_\allowbreak qaoa & 7 & 8 & 4.5 & 9 & \textbf{1} & 3 & 2 & 6 & 4.5 \\
ReactionOptimisation / mit\_\allowbreak case1\_\allowbreak mixed & 2.5 & 8 & 7 & \textbf{1} & 2.5 & 6 & 4 & 9 & 5 \\
ReactionOptimisation / reizman\_\allowbreak suzuki\_\allowbreak pareto & 3 & 6 & 7 & 4 & \textbf{1} & 2 & 5 & 9 & 8 \\
ReactionOptimisation / snar\_\allowbreak multiobjective & 2 & 8 & 6 & \textbf{1} & 7 & 4 & 3 & 9 & 5 \\
Robotics / DynamicObstacleAvoidanceNavigation & \textbf{1} & 8 & 5 & 2 & 9 & 3 & 4 & 7 & 6 \\
Robotics / PIDTuning & \textbf{1} & 9 & 5 & 6 & 8 & 4 & 7 & 2 & 3 \\
Robotics / QuadrupedGaitOptimization & 7 & 4 & 8.5 & 6 & \textbf{1} & 2 & 3 & 5 & 8.5 \\
Robotics / RobotArmCycleTimeOptimization & 4 & 9 & 8 & \textbf{1} & 6 & 3 & 6 & 6 & 2 \\
Robotics / UAVInspectionCoverageWithWind & 8.5 & 5 & 6 & 7 & 2 & 4 & 3 & \textbf{1} & 8.5 \\
SingleCellAnalysis / predict\_\allowbreak modality & 6 & 6 & 6 & \textbf{1} & 2 & 6 & 6 & 6 & 6 \\
StructuralOptimization / ISCSO2015 & \textbf{1} & 6 & 5 & 9 & 4 & 3 & 2 & 7 & 8 \\
StructuralOptimization / ISCSO2023 & \textbf{1} & 8 & 5 & 9 & 6 & 2 & 7 & 4 & 3 \\
StructuralOptimization / TopologyOptimization & 6 & 8 & 5 & 9 & \textbf{1} & 3 & 7 & 2 & 4 \\
SustainableDataCenterControl / hand\_\allowbreak written\_\allowbreak control & 3 & \textbf{1} & 2 & 8 & 9 & 4 & 5 & 6 & 7 \\
WirelessChannelSimulation / HighReliableSimulation & 2 & 5 & \textbf{1} & 9 & 4 & 6 & 3 & 7 & 8 \\
\midrule
Average & 3.63 & 6.71 & 5.86 & \textbf{3.54} & 4.81 & 4.34 & 4.76 & 5.82 & 5.53 \\
\bottomrule
\end{tabular}
\endgroup
\end{table*}

\begin{figure*}[t]
\centering
\begin{minipage}[t]{0.49\textwidth}
\centering
\includegraphics[width=\linewidth]{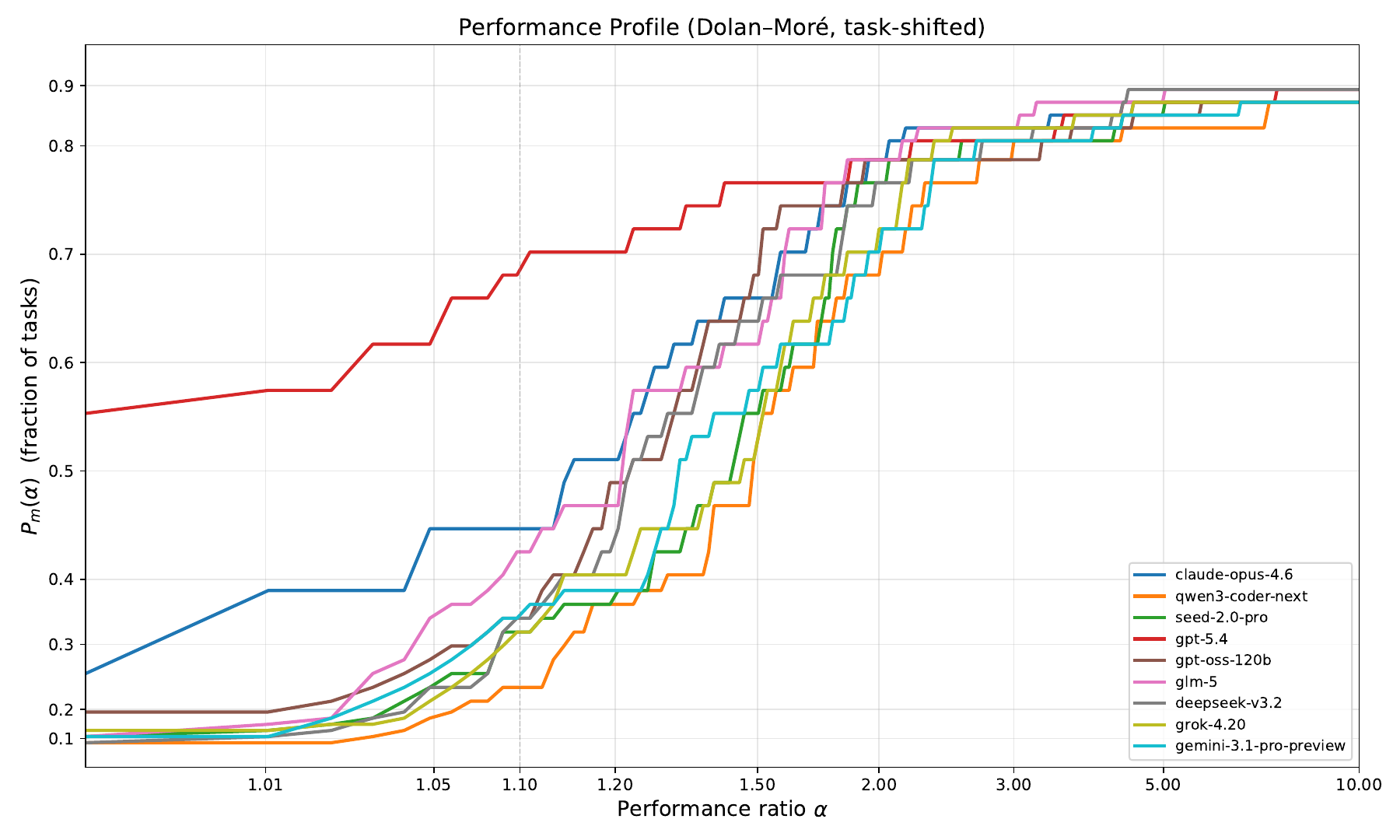}
\end{minipage}
\hfill
\begin{minipage}[t]{0.49\textwidth}
\centering
\includegraphics[width=\linewidth]{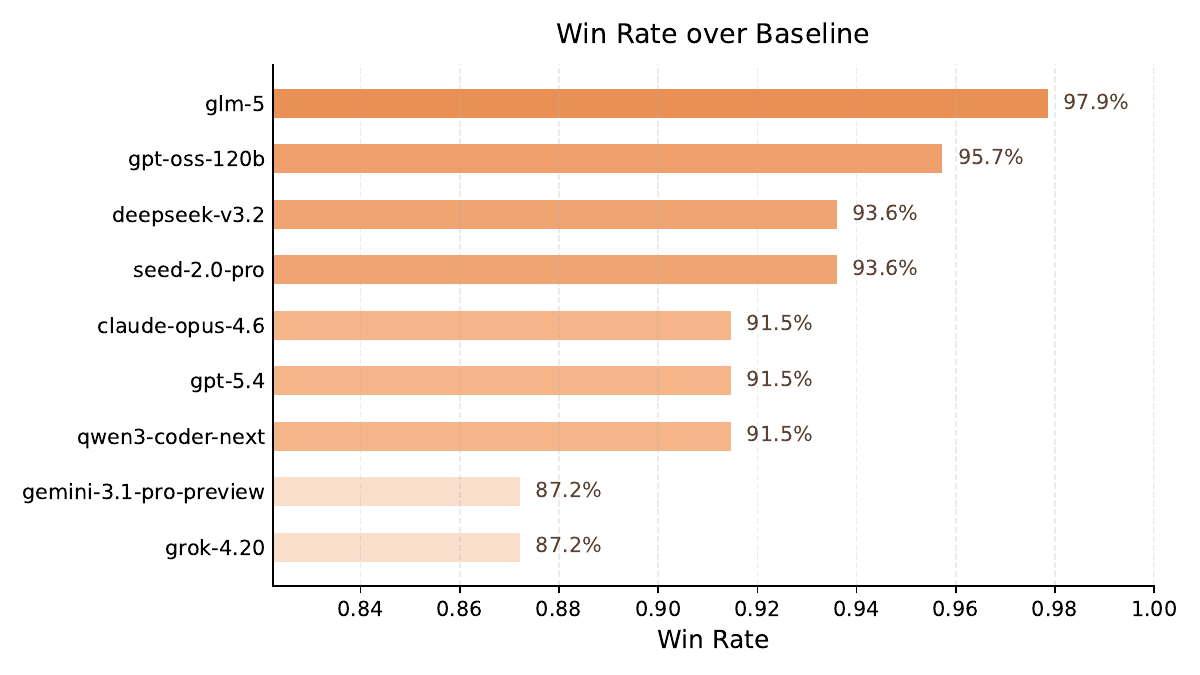}
\end{minipage}
\caption{Aggregate diagnostics of nine models on all 47 tasks under \texttt{openevolve}, after replacing \texttt{gpt-5.4} with the retest results. Left: Dolan--Mor\'e performance profile (higher and more left-shifted curves indicate better performance). Ratios are computed on per-task translated scores (see Section~\ref{sec:eval_protocol}): $\alpha=1.0$ means a model attains the task-best translated score (including ties), while $\alpha=1.1$ means the translated score is at least $1/1.1\approx90.9\%$ of the translated best. Under this definition, \texttt{gpt-5.4} is best on $24/47$ tasks ($51.1\%$), and is within the $1.1$ near-best band on $33/47$ tasks ($70.2\%$). Right: Win rate over baseline (fraction of tasks for which the model's best score exceeds the initial score).}
\label{fig:openevolve-profile-winrate-en}
\end{figure*}

Figure~\ref{fig:openevolve-profile-winrate-en} complements the task-level tables with two aggregate diagnostics. After the retest, the left panel (performance profile) shows \texttt{gpt-5.4} leading the strict near-best regime, while the right panel (win rate) continues to highlight \texttt{glm-5} and \texttt{gpt-oss-120b} as the most reliable models for improving over the baseline across the 47 tasks.

% TODO: Add Experiment 2: same FM, different evolving framework.

\subsection{Overall Comparison Between Models and Search Frameworks}

This section conducts a comparative study of \texttt{claude-opus-4.6} and \texttt{gpt-oss-120b} across three search frameworks from two complementary perspectives. At the aggregate level, we evaluate task-level outcomes using average ranks and win counts. At the trajectory level, we further analyze search statistics to understand how improvements are distributed over the course of optimization. Taken together, these results provide insight into both the relative effectiveness of the two models under each framework and the relationship between model capability and search dynamics.

\subsubsection{Aggregate Comparison}

% Table~\ref{tab:evolve-model-rank-by-framework-en} reports the average task rank of the two models under each of the three frameworks. Lower average rank is better.
Table~\ref{tab:evolve-model-rank-by-framework-en} reports the average within-task rank and win counts of the two models under each framework on the full 47-task rerun set. \texttt{claude-opus-4.6} outperforms \texttt{gpt-oss-120b} under all three frameworks, with lower average rank and more task wins in every case. The largest rank gap appears under \texttt{openevolve}, while \texttt{shinkaevolve} is close behind.
    
\newcommand{\bestnum}[1]{\textbf{\makebox[0pt][r]{\smash{\raisebox{0.65ex}{\scriptsize$\star$}}\hspace{0.18em}}#1}}

\begin{table}[t]
\centering
\scriptsize
\caption{Average ranks, wins, and ties of the two models under the three search frameworks on all tasks. We use Opus$=$\texttt{claude-opus-4.6} and OSS$=$\texttt{gpt-oss-120b}. A win is counted when one model's best score is strictly higher on a task; ties are listed separately. For rank aggregation, each task assigns rank 1 to the better model and rank 2 to the other model, while ties assign rank 1 to both models. Bold with $\star$ marks the better model in each row (lower is better for ranks; higher is better for wins).}
\label{tab:evolve-model-rank-by-framework-en}
\begin{tabular}{lcccccc}
\toprule
Framework & Opus Rank & OSS Rank & Opus Wins & OSS Wins & Ties \\
\midrule
\texttt{abmcts} & \bestnum{1.43} & 1.55 & \bestnum{26} & 20 & 1 \\
\texttt{openevolve} & \bestnum{1.28} & 1.64 & \bestnum{30} & 13 & 4 \\
\texttt{shinkaevolve} & \bestnum{1.28} & 1.62 & \bestnum{29} & 13 & 5 \\
\bottomrule
\end{tabular}
\end{table}

At the same time, the relative closeness of \texttt{openevolve} and \texttt{shinkaevolve} within each model indicates that framework quality is not absolute. Rather, the results point to a model--framework interaction: different search mechanisms expose different strengths of the underlying model, even when the overall ranking between models remains stable.

\subsubsection{Trajectory Tendencies Across Frameworks}
\label{sec:trajectory-tendencies}

Table~\ref{tab:evolve-trajectory-tendencies-en} reports median trajectory statistics to distinguish early aggressive rewrites from sustained refinement. Concretely, \emph{gain} is the median baseline-relative percentage improvement; \emph{early-share} is the fraction of total improvement obtained in the first quarter of the trajectory; \emph{last-improve} is the normalized position of the final improvement event (larger means later); \emph{best-updates} is the number of running-best updates; and \emph{plateau} is the longest no-improvement segment as a fraction of total steps (smaller means fewer stagnation steps). We use these metrics to compare the two models in terms of early-gain concentration, late-stage updates, and plateau behavior.

\begin{table*}[t]
\centering
\scriptsize
\caption{Trajectory tendency statistics for each model--framework combination, with Opus$=$\texttt{claude-opus-4.6} and OSS$=$\texttt{gpt-oss-120b}. Metrics are defined as: \emph{gain} (median baseline-relative improvement), \emph{early-share} (fraction of total improvement achieved in the first quarter), \emph{last-improve} (normalized step index of the final improvement; larger is later), \emph{best-updates} (count of running-best updates), and \emph{plateau} (longest no-improvement fraction). Bold with $\star$ marks the best value in each metric column (higher is better for gain, early-share, last-improve, and best-updates; lower is better for plateau).}
\label{tab:evolve-trajectory-tendencies-en}
\begin{tabular}{lccccc}
\toprule
Model + Framework & gain(\%) & early-share & last-improve & best-updates & plateau \\
\midrule
Opus + \texttt{abmcts} & 61.6 & 0.89 & 0.53 & 4.0 & 0.65 \\
Opus + \texttt{openevolve} & \bestnum{73.3} & 0.97 & \bestnum{0.70} & 5.5 & 0.51 \\
Opus + \texttt{shinkaevolve} & 72.9 & 0.95 & 0.44 & 4.0 & 0.62 \\
OSS + \texttt{abmcts} & 50.4 & \bestnum{0.99} & 0.30 & 3.0 & 0.72 \\
OSS + \texttt{openevolve} & 58.9 & \bestnum{0.99} & 0.53 & \bestnum{6.0} & 0.57 \\
OSS + \texttt{shinkaevolve} & 68.3 & 0.89 & 0.64 & 5.0 & \bestnum{0.45} \\
\bottomrule
\end{tabular}
\end{table*}

\paragraph{ABMCTS: gains concentrate early, with limited late updates.}
ABMCTS places most gains in early iterations for both models (\emph{early-share} $=0.89$ on \texttt{claude-opus-4.6} and $0.99$ on \texttt{gpt-oss-120b}). \texttt{gpt-oss-120b} also shows a long plateau ($0.72$) and only $3$ median \emph{best-updates}, indicating fewer late improvements. Under this setting, \texttt{claude-opus-4.6} has better aggregate rank and win statistics.

\paragraph{OpenEvolve: early concentration with continued refinement.}
OpenEvolve also concentrates gains early (\emph{early-share} $=0.97$ on \texttt{claude-opus-4.6} and $0.99$ on \texttt{gpt-oss-120b}), while still permitting multiple later improvements (\emph{best-updates} $=5.5$ and $6.0$, respectively). Compared with ABMCTS, this setting leaves more room for late-stage updates.

\paragraph{ShinkaEvolve: later improvements and shorter plateaus.}
ShinkaEvolve shows more sustained late-stage updates, especially for \texttt{gpt-oss-120b} (\emph{last-improve} $=0.64$, \emph{plateau} $=0.45$, \emph{best-updates} $=5$). \texttt{claude-opus-4.6} under \texttt{shinkaevolve} retains high median \emph{gain} ($72.9\%$) with strong early progress (\emph{early-share} $=0.95$). Overall, both models exhibit longer effective refinement than under ABMCTS.

\subsubsection{Model Behavioral Differences}
 
The trajectory statistics above provide a model-level comparison beyond average rank.
 
\paragraph{\texttt{claude-opus-4.6}: stronger early-result quality.} Within this two-model comparison, \texttt{claude-opus-4.6} achieves lower average within-task rank than \texttt{gpt-oss-120b} across all three frameworks. The gap is largest under ABMCTS, where outcomes are more sensitive to early accepted candidates, and remains present under OpenEvolve and ShinkaEvolve.
 
\paragraph{\texttt{gpt-oss-120b}: more persistent late-stage updates.} Under OpenEvolve and ShinkaEvolve, \texttt{gpt-oss-120b} has a higher median \emph{best-updates} count than \texttt{claude-opus-4.6}, with shorter plateau durations. This indicates more frequent late-stage improvements.
 
\paragraph{The same framework yields different trajectories by model.} Under ShinkaEvolve, \texttt{claude-opus-4.6} reaches its final improvement earlier, while \texttt{gpt-oss-120b} tends to improve later. Accordingly, the model gap is smaller on refinement-heavy tasks and larger on tasks where early high-quality candidates dominate.

% \subsubsection{Conclusion}

% The experiments show that the three search frameworks exhibit clearly different behavioral modes on the two models. \texttt{abmcts} is more conservative and more prone to early lock-in; \texttt{openevolve} is more aggressive and more dependent on early structural rewrites; \texttt{shinkaevolve} is closer to a controlled sustained-refinement mechanism. Claude performs better under all three frameworks, which suggests stronger candidate quality overall; GPT-OSS, however, displays greater path plasticity under \texttt{openevolve} and \texttt{shinkaevolve}, making it more capable of producing continued improvements later in the trajectory.

% The results therefore do not support the view that any one framework or model is universally superior. Rather, they point to a strong interaction between model capability and search mechanism. The key to code evolution is not simply to search more, but to determine how aggressively to allocate effort among early rewrites, local repair, and late refinement; the model difference is visible precisely in how well each model supports these different behavioral regimes.

\subsubsection{Summary}

The three frameworks occupy distinct positions along the exploration--exploitation spectrum. Specifically, \texttt{abmcts} is the most conservative, \texttt{openevolve} is the most structurally aggressive, and \texttt{shinkaevolve} is the most conducive to sustained refinement. Within this two-model comparison, the advantage of \texttt{claude-opus-4.6} over \texttt{gpt-oss-120b} is most pronounced on tasks where early accepted candidates exert a strong influence on final outcomes, whereas \texttt{gpt-oss-120b} reduces this gap on tasks that permit continued gains through late-stage updates. These findings suggest that framework choice should be informed not only by aggregate model performance, but also by model-specific trajectory characteristics.

\subsection{Optimization Dynamics}
\label{sec:opt_dynamics}

% ─────────────────────────────────────────────────────────────────────────────

\subsubsection{Improvement Dynamics Follow a Power Law}
\label{sec:long_horizon}

We run \textsc{OpenEvolve} with GPT-OSS-120B on all 47 tasks for 500
iterations, tracking each iteration at which the running best score improves.

\begin{figure}[t]
  \centering
  \includegraphics[width=\linewidth]{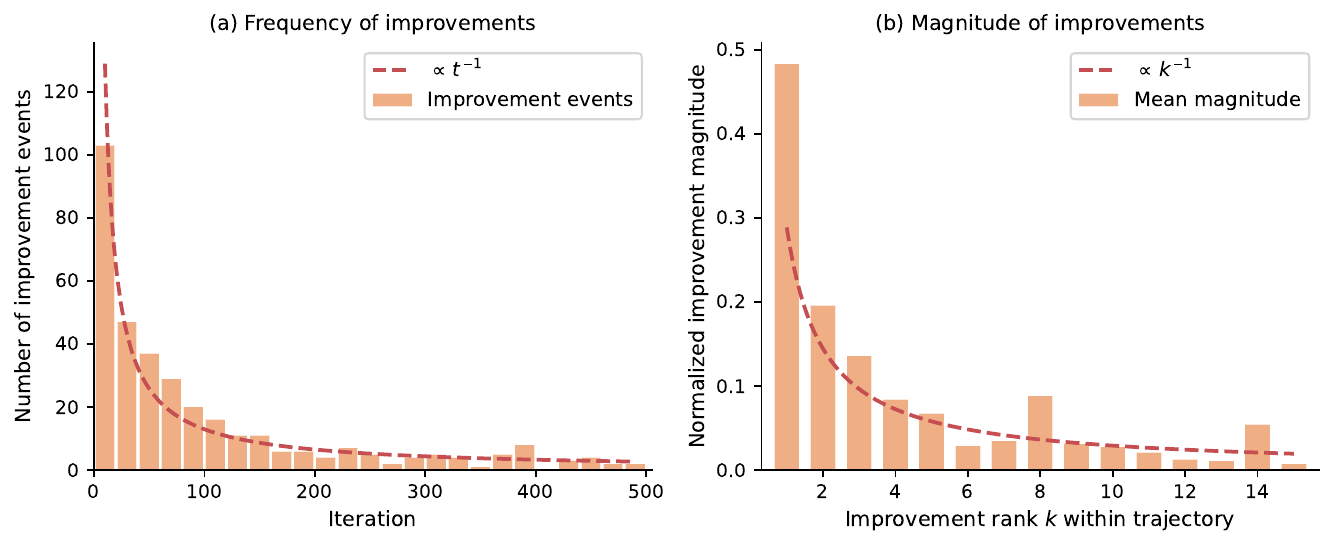}
  \caption{
    \textbf{Dual $-1$ power-law decay in improvement dynamics across
    47 tasks (GPT-OSS-120B, \textsc{OpenEvolve}, 500 iterations).}
    \textbf{(a)} Histogram of improvement events by iteration; dashed line
    is a $\propto t^{-1}$ fit ($R^2{=}0.84$).
    \textbf{(b)} Median normalized improvement magnitude by improvement rank
    $k$ within each task's trajectory; dashed line is a $\propto k^{-1}$ fit,
    showing that improvement size shrinks with ordinal rank just as improvement
    frequency shrinks with iteration.
  }
  \label{fig:longhorizon}
\end{figure}

\paragraph{Results.}
Figure~\ref{fig:longhorizon} reveals a \emph{dual power-law} structure.
Panel~(a) shows that the frequency of improvement events decays as
$\propto t^{-1}$ with iteration: the majority occur in the first $\sim$30
steps, with a long tail to iteration 500.
Panel~(b) shows that the \emph{magnitude} of the $k$-th improvement within
each task's trajectory obeys the same $\propto k^{-1}$ law: the first
improvement is a large structural rewrite, while each subsequent one is
a smaller incremental refinement.
Both fits hold with $R^2 > 0.83$ under a constrained slope of $-1$.
All 47 tasks record at least one improvement; the median number of
improvements per task is 7.

\paragraph{Takeaway.}
Improvements become both rarer and smaller over time: frequency decays as
$t^{-1}$, and per-improvement magnitude decays as $k^{-1}$. The result is a
double squeeze that quickly drives marginal returns near zero after
$\sim$50--100 iterations. This motivates the budget analysis in
Section~\ref{sec:depth_width}.

% ─────────────────────────────────────────────────────────────────────────────
\subsubsection{Depth Dominates Width under Fixed Budget}
\label{sec:depth_width}

The power-law structure above raises a natural question: given a fixed
evaluation budget $B$, is it better to run a single deep chain of $B$ steps,
or to spread the budget across $n$ independent chains of depth $d = B/n$ and
take the best?

We answer this by fixing the total budget $B = n \times d \le 256$ and varying
$n \in \{1, 2, 4, 8, 16\}$ on a 10-task subset. For each task, we compute the
best-of-$n$ score for every $(n, d)$ configuration, then normalize by the
maximum across all configurations; thus $1.0$ indicates the best observed
performance in this experiment.

\begin{figure}[t]
  \centering
  \includegraphics[width=0.52\linewidth]{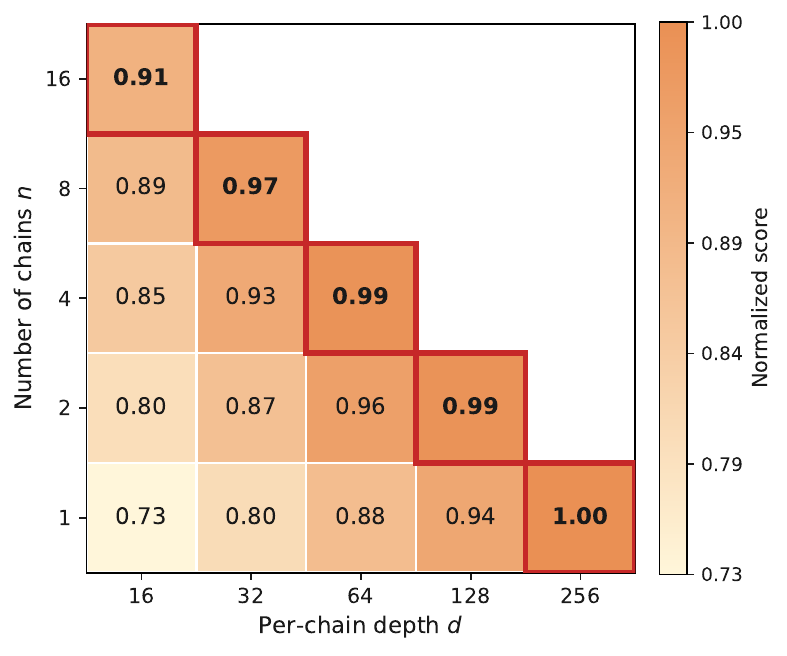}
  \caption{
    \textbf{Normalized best-of-$n$ score across depth and width configurations
    (GPT-OSS-120B, \textsc{OpenEvolve}).}
    Each cell shows the mean normalized score for $n$ chains each run to depth $d$.
    Scores are normalized per task by the maximum across all configurations.
    Red borders mark $n \times d = 256$: scores decrease with $n$, showing depth dominates width.
  }
  \label{fig:ablation_heatmap}
\end{figure}

\paragraph{Results.}
Figure~\ref{fig:ablation_heatmap} shows the normalized best-of-$n$ score for
all $(n, d)$ pairs. Along the equal-budget diagonal ($n \times d = 256$, red
border), the score decreases monotonically with $n$: $1.00$, $0.99$,$0.99$, $0.97$,
$0.91$ for $n = 1, 2, 4, 8, 16$,
confirming that \emph{depth matters more than width} under a fixed budget.
Width does help when depth is held fixed (scores increase upward within each
column), but under a fixed total budget, the single deep chain dominates.

\paragraph{Takeaway.}
Iteration depth correlates strongly with solution quality: a generative
optimization agent must pursue a single reasoning chain far enough for
structural breakthroughs. Restarting resets accumulated context, distinguishing
generative optimization from population-based methods.

\section{Related work}
We group prior work into three strands that contribute ingredients of our setting: agent benchmarks for open-world problem solving, engineering-grounded benchmarks and systems, and optimization methods built around verifier-guided search. The key distinction for Frontier-Eng is not any one ingredient in isolation, but their combination in a cross-domain benchmark with executable constraints and limited interaction budgets.

\begin{table}[t]
\caption{Selected adjacent benchmarks and the comparison dimensions used to position Frontier-Eng. \yes, \no, and \partialyes\ denote that a property is fully present, absent, or partially present (e.g., for some tasks or only indirectly emphasized). ``Hard constraints'' refers to explicit valid/invalid conditions beyond answer accuracy alone. ``Best-found under budget'' refers to evaluation by the best feasible solution found within a fixed interaction budget.}
\label{tab:related_comparison}
\centering
\scriptsize
\resizebox{.8\linewidth}{!}{%
\begin{tabular}{@{}p{6.4cm}ccccc@{}}
\toprule
Benchmark & \shortstack{Interactive\\env.} & \shortstack{Hard\\constraints} & \shortstack{Exec.\\verifier} & \shortstack{Cross-\\domain} & \shortstack{Best-found\\under budget} \\
\midrule
\multicolumn{6}{@{}l}{\textbf{Engineering QA benchmarks}} \\
Transportation systems \citep{Syed2024Benchmarking} & \no & \no & \no & \no & \no \\
Control engineering \citep{Kevian2024CapabilitiesOL} & \no & \no & \no & \no & \no \\
EEE-Bench \citep{Li2024EEEBenchAC} & \no & \no & \no & \no & \no \\
CIRCUIT \citep{Skelic2025CIRCUITAB} & \no & \no & \no & \no & \no \\
\midrule
\multicolumn{6}{@{}l}{\textbf{Open-world agent benchmarks}} \\
SWE-bench \citep{Jimenez2023SWEbenchCL} & \yes & \no & \yes & \no & \no \\
MLE-bench \citep{Chan2024MLEbenchEM} & \yes & \no & \yes & \partialyes & \no \\
RE-Bench \citep{Wijk2024REBenchEF} & \yes & \no & \partialyes & \partialyes & \no \\
PaperBench \citep{Starace2025PaperBenchEA} & \yes & \no & \yes & \no & \no \\
Terminal-Bench \citep{Merrill2026TerminalBenchBA} & \yes & \no & \yes & \partialyes & \no \\
The AI Scientist \citep{Lu2024TheAS} & \yes & \no & \yes & \no & \no \\
AlphaEvolve \citep{Novikov2025AlphaEvolveAC} & \yes & \yes & \yes & \yes & \yes \\
\midrule
\multicolumn{6}{@{}l}{\textbf{Engineering design benchmarks}} \\
EngDesign \citep{Guo2025EngineeringAGI} & \partialyes & \yes & \yes & \yes & \partialyes \\
EngiBench \citep{Zhou2025EngiBenchAB} & \partialyes & \partialyes & \partialyes & \yes & \partialyes \\
BikeBench \citep{Regenwetter2025BikeBenchAB} & \no & \yes & \yes & \no & \partialyes \\
BuildArena \citep{Xia2025BuildArenaAP} & \partialyes & \yes & \yes & \no & \partialyes \\
\midrule
\multicolumn{6}{@{}l}{\textbf{This work}} \\
Frontier-Eng (Ours) & \yes & \yes & \yes & \yes & \yes \\
\bottomrule
\end{tabular}
}
\end{table}

\textbf{Agent benchmarks for open-world problem solving.}
A growing body of work evaluates AI agents on long-horizon tasks in realistic environments rather than short, self-contained problems. AgentBench evaluates general-purpose LLM agents across multiple interactive environments \citep{Liu2024AgentBench}, while SWE-bench studies whether language models can resolve real GitHub issues in existing repositories \citep{Jimenez2023SWEbenchCL}. MLE-bench extends this paradigm to end-to-end machine learning engineering \citep{Chan2024MLEbenchEM}. RE-Bench and PaperBench examine frontier R\&D workflows and paper reproduction \citep{Wijk2024REBenchEF,Starace2025PaperBenchEA}, while WebArena, BrowseComp, OSWorld, and Terminal-Bench evaluate tool use in realistic web, computer-use, and command-line settings \citep{Zhou2024WebArena,Wei2025BrowseCompAS,Xie2024OSWorld,Merrill2026TerminalBenchBA}. These benchmarks establish important protocols for planning, tool use, and iterative correction in open environments.

At the same time, their task environments remain primarily digital: codebases, datasets, documents, and logs. Success is usually defined by task completion or artifact correctness in software and research workflows, rather than by how effectively an agent improves a feasible engineering design under simulator feedback, hard constraints, and a fixed search budget. Frontier-Eng adopts the long-horizon, tool-using agent perspective of this literature, but grounds evaluation in executable engineering tasks whose quality depends on constrained optimization rather than solely on digital task completion.

\textbf{Engineering-grounded benchmarks and systems.}
Engineering-focused evaluation has developed along two complementary directions. The first measures engineering knowledge and domain reasoning through question answering or structured problem solving in areas such as transportation systems, control engineering, additive manufacturing, and electrical engineering \citep{Syed2024Benchmarking,Kevian2024CapabilitiesOL,Eslaminia2024FDMBench,Li2024EEEBenchAC,Skelic2025CIRCUITAB,Chen2025BenchmarkingLL}. These benchmarks are valuable for testing technical knowledge, but they remain largely answer-centric and only partially capture the iterative, tool-grounded character of real design workflows.

The second direction moves closer to executable engineering evaluation. \textsc{EngDesign} uses simulation-based verification for multi-domain design tasks \citep{Guo2025EngineeringAGI}. EngiBench studies engineering problem solving and open-ended modeling \citep{Zhou2025EngiBenchAB}. Domain-specific efforts such as BikeBench and BuildArena similarly emphasize grounded constraints and executable feedback \citep{Regenwetter2025BikeBenchAB,Xia2025BuildArenaAP}. In parallel, systems such as ControlAgent, AnalogCoder, and SPICED demonstrate the value of coupling LLMs with engineering tools and verifiers in specific domains \citep{Guo2024ControlAgentAC,Lai2024AnalogCoderAC,Chaudhuri2024SPICED}. As summarized in Table~\ref{tab:related_comparison}, Frontier-Eng builds on this trend but places more emphasis on benchmark-scale coverage across heterogeneous domains, a shared agent-facing protocol, and a primary evaluation target based on budget-aware best-found feasible performance rather than only validity or task completion.

\textbf{Verifier-guided search and evaluation.}
Frontier-Eng is also motivated by work that treats LLMs as components in iterative optimization loops. ReAct established interleaved reasoning and acting for tool-using language agents \citep{Yao2023ReAct}, Reflexion introduced feedback-driven self-improvement through verbal reinforcement \citep{Shinn2023Reflexion}, Tree of Thoughts framed deliberate inference-time search over multiple reasoning paths \citep{yao2023tree}, and LATS connected these ideas to explicit tree search for language agents \citep{Zhou2024LATS}. Within verifier-guided optimization more specifically, OPRO showed that language models can improve candidate solutions using feedback \citep{Yang2023LargeLM}, FunSearch demonstrated that executable program search guided by evaluation can yield mathematical discoveries \citep{DBLP:journals/nature/RomeraParedesBNBKDREWFKF24}, The AI Scientist \citep{Lu2024TheAS} introduced a system for fully automated scientific discovery, AlphaEvolve \citep{Novikov2025AlphaEvolveAC} demonstrated evolutionary code editing for algorithmic discovery, Evolution of Heuristics extended this perspective to automatic heuristic and algorithm design \citep{Liu2024EvolutionOH}, and iterative refinement methods such as Self-Refine likewise emphasize repeated propose--evaluate--revise cycles \citep{Madaan2023SelfRefineIR}. Across these efforts, the core idea is that evaluation is most informative when intermediate proposals can be checked and improved rather than judged only once.

This perspective is especially natural for engineering, where simulators, solvers, and rule-based verifiers can provide grounded signals about feasibility and performance. Recent discussions of engineering intelligence similarly argue that evaluation should focus on executable artifacts, tool use, structured outputs, and constraints grounded in real systems rather than on text-only answers \citep{Neema2025OnTE}. Frontier-Eng operationalizes this view as a benchmark protocol: under a fixed interaction budget, the central question is not whether an agent produces a single correct response, but how effectively it searches for high-quality feasible solutions across diverse engineering domains.

\section{Conclusion}

We introduced \textbf{Frontier-Eng}, a large-scale benchmark for evaluating AI agents on generative optimization tasks across five broad engineering categories. By formalizing engineering optimization as a triple of context, initial solution, and executable evaluator, we shifted the evaluation focus from binary pass/fail outcomes to the iterative, budget-aware search for high-quality feasible solutions. Our benchmark packages $47$ tasks with unified metadata-driven interfaces, grounded in industrial-grade simulators and solvers that provide reliable feedback while enforcing strict anti-hack safeguards.

The evaluation of frontier language models and representative optimization strategies reveals both an emerging capacity for iterative improvement and significant remaining challenges. While current agents can effectively optimize simple controllers and algorithmic kernels, they often struggle with the multi-objective trade-offs and long-horizon reasoning required by complex physical systems. These findings suggest that the path toward engineering artificial general intelligence lies not only in larger models but in the development of search strategies that can more effectively integrate domain knowledge with structured feedback from executable environments.

By providing a stable, cross-disciplinary platform for comparing generative optimization methods, Frontier-Eng aims to accelerate research into agents that can participate in the core iterative cycle of engineering: define, build, test, and refine. As we expand the benchmark to include more diverse domains and higher-fidelity simulations, we hope it serves as a catalyst for developing AI systems that deliver tangible value through the systematic optimization of real-world engineering systems.

\clearpage
\newpage
\definecolor{brandcolor}{RGB}{227, 170, 121}
\section{Contributions and Acknowledgements}
The contributors’ names are sorted in alphabetical order of the last name.

\textbf{Project Lead}

Dapeng Jiang (\textcolor{brandcolor}{jdp22@mails.tsinghua.edu.cn})

\textbf{Core Contributors}

Yizhe Chi, Deyao Hong, Dapeng Jiang, Tianwei Luo, Qinhuai Na, Kaisen Yang, Boshi Zhang

\textbf{Contributors}

Zhe Cao, Xiaoyan Fan, Bingxiang He, Han Hao, Weiyang Jin, Dianqiao Lei, Qingle Liu, Houde Qian, Bowen Wang, Situ Wang, Youjie Zheng, Yifan Zhou

\textbf{Team Management}

Calvin Xiao, Eren Cai, Qinhuai Na

\textbf{Corresponding to} Qinhuai Na (\textcolor{brandcolor}{nana@einsia.ai}).

\textbf{Acknowledgements.} This project is founded and supported by Navers Lab, Einsia.AI.

% ── References ──────────────────────────────────────────────────────────────

\clearpage
\newpage
\bibliographystyle{assets/plainnat}
\bibliography{ref}

@Article{Zhou2025EngiBenchAB,
 author = {Xiyuan Zhou and others},
 journal = {ArXiv},
 title = {EngiBench: A Benchmark for Evaluating Large Language Models on Engineering Problem Solving},
 volume = {abs/2509.17677},
 year = {2025}
}

@Article{Regenwetter2025BikeBenchAB,
 author = {Lyle Regenwetter and others},
 journal = {ArXiv},
 title = {Bike-Bench: A Bicycle Design Benchmark for Generative Models with Objectives and Constraints},
 volume = {abs/2508.00830},
 year = {2025}
}

@Article{Xia2025BuildArenaAP,
 author = {Tian Xia and others},
 journal = {ArXiv},
 title = {BuildArena: A Physics-Aligned Interactive Benchmark of LLMs for Engineering Construction},
 volume = {abs/2510.16559},
 year = {2025}
}

@Article{Guo2024ControlAgentAC,
 author = {Xing-ming Guo and others},
 journal = {ArXiv},
 title = {ControlAgent: Automating Control System Design via Novel Integration of LLM Agents and Domain Expertise},
 volume = {abs/2410.19811},
 year = {2024}
}

@Article{Lai2024AnalogCoderAC,
 author = {Yao Lai and others},
 journal = {ArXiv},
 title = {AnalogCoder: Analog Circuit Design via Training-Free Code Generation},
 volume = {abs/2405.14918},
 year = {2024}
}

@Article{Kevian2024CapabilitiesOL,
 author = {Darioush Kevian and others},
 journal = {ArXiv},
 title = {Capabilities of Large Language Models in Control Engineering: A Benchmark Study on GPT-4, Claude 3 Opus, and Gemini 1.0 Ultra},
 volume = {abs/2404.03647},
 year = {2024}
}

@Article{Eslaminia2024FDMBench,
 author = {Ahmadreza Eslaminia and others},
 journal = {ArXiv},
 title = {FDM-Bench: A Comprehensive Benchmark for Evaluating Large Language Models in Additive Manufacturing Tasks},
 volume = {abs/2412.09819},
 year = {2024}
}

@Article{Syed2024Benchmarking,
 author = {U. Syed and others},
 journal = {ArXiv},
 title = {Benchmarking the Capabilities of Large Language Models in Transportation System Engineering},
 volume = {abs/2408.08302},
 year = {2024}
}

@article{DBLP:journals/nature/RomeraParedesBNBKDREWFKF24,
  author       = {Bernardino Romera{-}Paredes and
                  Mohammadamin Barekatain and
                  Alexander Novikov and
                  Matej Balog and
                  M. Pawan Kumar and
                  Emilien Dupont and
                  Francisco J. R. Ruiz and
                  Jordan S. Ellenberg and
                  Pengming Wang and
                  Omar Fawzi and
                  Pushmeet Kohli and
                  Alhussein Fawzi},
  title        = {Mathematical discoveries from program search with large language models},
  journal      = {Nat.},
  volume       = {625},
  number       = {7995},
  pages        = {468--475},
  year         = {2024},
  url          = {https://doi.org/10.1038/s41586-023-06924-6},
  doi          = {10.1038/S41586-023-06924-6},
  bibsource    = {dblp computer science bibliography, https://dblp.org}
}

@Article{Jain2024LiveCodeBenchHA,
 author = {Naman Jain and King Han and Alex Gu and Wen-Ding Li and Fanjia Yan and Tianjun Zhang and Sida Wang and Armando Solar-Lezama and Koushik Sen and Ion Stoica},
 booktitle = {International Conference on Learning Representations},
 journal = {ArXiv},
 title = {LiveCodeBench: Holistic and Contamination Free Evaluation of Large Language Models for Code},
 volume = {abs/2403.07974},
 year = {2024}
}

@Article{Skelic2025CIRCUITAB,
 author = {Lejla Skelic and Yan Xu and Matthew B. Cox and Wenjie Lu and Tao Yu and R. Han},
 booktitle = {arXiv.org},
 journal = {ArXiv},
 title = {CIRCUIT: A Benchmark for Circuit Interpretation and Reasoning Capabilities of LLMs},
 volume = {abs/2502.07980},
 year = {2025}
}

@Book{Blockley2012EngineeringAV,
 author = {D. Blockley},
 title = {Engineering: A Very Short Introduction},
 publisher = {Oxford University Press},
 year = {2012}
}

@Book{Grote2009SpringerHO,
 author = {Karl-Heinrich Grote and E. Antonsson},
 title = {Springer handbook of mechanical engineering},
 publisher = {Springer},
 year = {2009}
}

@Book{Chen2002TheCE,
 author = {Wai-Fah Chen and J. Liew},
 title = {The civil engineering handbook},
 publisher = {CRC Press},
 year = {2002}
}

@Book{Chen2004TheEE,
 author = {Wai-Kai Chen},
 title = {The Electrical Engineering Handbook},
 publisher = {Elsevier},
 year = {2004}
}

@Article{Yang2023LargeLM,
 author = {Chengrun Yang and Xuezhi Wang and Yifeng Lu and Hanxiao Liu and Quoc V. Le and Denny Zhou and Xinyun Chen},
 journal = {arXiv preprint arXiv:2309.03409},
 title = {Large Language Models as Optimizers},
 year = {2023}
}

@InProceedings{Liu2024EvolutionOH,
 author = {Fei Liu and Xialiang Tong and Mingxuan Yuan and Xi Lin and Fu Luo and Zhenkun Wang and Zhichao Lu and Qingfu Zhang},
 booktitle = {International Conference on Machine Learning},
 pages = {32201-32223},
 title = {Evolution of Heuristics: Towards Efficient Automatic Algorithm Design Using Large Language Model},
 year = {2024}
}

@Article{Madaan2023SelfRefineIR,
 author = {Aman Madaan and Niket Tandon and Prakhar Gupta and Skyler Hallinan and Luyu Gao and Sarah Wiegreffe and Uri Alon and Nouha Dziri and Shrimai Prabhumoye and Yiming Yang and S. Welleck and Bodhisattwa Prasad Majumder and Shashank Gupta and A. Yazdanbakhsh and Peter Clark},
 journal = {arXiv preprint arXiv:2303.17651},
 title = {Self-Refine: Iterative Refinement with Self-Feedback},
 year = {2023}
}

@Article{Jimenez2023SWEbenchCL,
 author = {Carlos E. Jimenez and John Yang and Alexander Wettig and Shunyu Yao and Kexin Pei and Ofir Press and Karthik Narasimhan},
 journal = {arXiv preprint arXiv:2310.06770},
 title = {SWE-bench: Can Language Models Resolve Real-World GitHub Issues?},
 year = {2023}
}

@Article{Chan2024MLEbenchEM,
 author = {Jun Shern Chan and Neil Chowdhury and Oliver Jaffe and J. Aung and Dane Sherburn and E. Mays and Giulio Starace and Kevin Liu and Leon Maksin and Tejal Patwardhan and Lilian Weng and Aleksander Mkadry},
 booktitle = {arXiv.org},
 journal = {ArXiv},
 title = {MLE-bench: Evaluating Machine Learning Agents on Machine Learning Engineering},
 volume = {abs/2410.07095},
 year = {2024}
}

@Article{Starace2025PaperBenchEA,
 author = {Giulio Starace and Oliver Jaffe and Dane Sherburn and J. Aung and Jun Shern Chan and Leon Maksin and Rachel Dias and E. Mays and Benjamin Kinsella and Wyatt Thompson and Johannes Heidecke and Amelia Glaese and Tejal Patwardhan},
 booktitle = {International Conference on Machine Learning},
 journal = {ArXiv},
 title = {PaperBench: Evaluating AI's Ability to Replicate AI Research},
 volume = {abs/2504.01848},
 year = {2025}
}

@Article{Wijk2024REBenchEF,
 author = {Hjalmar Wijk and T. Lin and Joel Becker and Sami Jawhar and Neev Parikh and Thomas Broadley and Lawrence Chan and Michael Chen and Joshua Clymer and Jai Dhyani and Elena Ericheva and Katharyn Garcia and Brian Goodrich and Nikola Jurkovic and Megan Kinniment and Aron Lajko and Seraphina Nix and L. Sato and William Saunders and M. Taran and Ben West and Elizabeth Barnes},
 booktitle = {arXiv.org},
 journal = {ArXiv},
 title = {RE-Bench: Evaluating frontier AI R\&D capabilities of language model agents against human experts},
 volume = {abs/2411.15114},
 year = {2024}
}

@Article{Naug2024SustainDCB,
 author = {Avisek Naug and Antonio Guillen and Ricardo Luna and Vineet Gundecha and Desik Rengarajan and Sahand Ghorbanpour and Sajad Mousavi and Ashwin Ramesh Babu and Dejan Markovikj and L. D. Kashyap and S. Sarkar},
 booktitle = {arXiv.org},
 journal = {ArXiv},
 title = {SustainDC - Benchmarking for Sustainable Data Center Control},
 volume = {abs/2408.07841},
 year = {2024}
}

@Article{Lu2024TheAS,
 author = {Chris Lu and Cong Lu and R. Lange and J. Foerster and Jeff Clune and David Ha},
 booktitle = {arXiv.org},
 journal = {ArXiv},
 title = {The AI Scientist: Towards Fully Automated Open-Ended Scientific Discovery},
 volume = {abs/2408.06292},
 year = {2024}
}

@Article{Novikov2025AlphaEvolveAC,
 author = {Alexander Novikov and Ngân V˜u and Marvin Eisenberger and Emilien Dupont and Po-Sen Huang and Adam Zsolt Wagner and S. Shirobokov and Borislav M. Kozlovskii and Francisco J. R. Ruiz and Abbas Mehrabian and M. P. Kumar and Abigail See and Swarat Chaudhuri and George Holland and A. Davies and Sebastian Nowozin and Pushmeet Kohli and Matej Balog and Google DeepMind},
 booktitle = {arXiv.org},
 journal = {ArXiv},
 title = {AlphaEvolve: A coding agent for scientific and algorithmic discovery},
 volume = {abs/2506.13131},
 year = {2025}
}

@Article{Wei2025BrowseCompAS,
 author = {Jason Wei and Zhiqing Sun and Spencer Papay and Scott McKinney and Jeff Han and Isa Fulford and Hyung Won Chung and Alexandre Passos and W. Fedus and Amelia Glaese},
 booktitle = {arXiv.org},
 journal = {ArXiv},
 title = {BrowseComp: A Simple Yet Challenging Benchmark for Browsing Agents},
 volume = {abs/2504.12516},
 year = {2025}
}

@Article{Neema2025OnTE,
 author = {Sandeep Neema and Susmit Jha and Adam Nagel and Ethan Lew and C. Sureshkumar and Aleksa Gordic and C. Shimmin and Hieu Nguygen and P. Eremenko},
 booktitle = {arXiv.org},
 journal = {ArXiv},
 title = {On the Evaluation of Engineering Artificial General Intelligence},
 volume = {abs/2505.10653},
 year = {2025}
}

@Article{Merrill2026TerminalBenchBA,
 author = {Mike A. Merrill and Alexander G Shaw and Nicholas Carlini and Boxuan Li and Harsh Raj and Ivan Bercovich and Lin Shi and J. Shin and Thomas Walshe and E. K. Buchanan and Junhong Shen and Guanghao Ye and Hao Lin and Jason Poulos and Maoyu Wang and Marianna Nezhurina and J. Jitsev and Di Lu and O. M. Mastromichalakis and Zhiwei Xu and Zizhao Chen and Yue Liu and Robert Zhang and L. Chen and Anurag Kashyap and Jan-Lucas Uslu and Jeffrey Li and Jianbo Wu and Minghao Yan and Song Bian and Vedang Sharma and Ke Sun and Steven Dillmann and Akshay Anand and Andrew Lanpouthakoun and Bardia Koopah and Changran Hu and E. Guha and Gabriel H. S. Dreiman and Jiacheng Zhu and Karl Krauth and Li Zhong and Niklas Muennighoff and Robert K. Amanfu and Shangyin Tan and Shreyas Pimpalgaonkar and Tushar Aggarwal and Xia Lin and Xin Lan and Xuandong Zhao and Yiqing Liang and Yuanli Wang and Zilong Wang and Changzhi Zhou and David Heineman and Hange Liu and H. Trivedi and John Yang and Junhong Lin and Manish Shetty and Michael Yang and Nabil Omi and Negin Raoof and Shanda Li and Terry Yue Zhuo and Wu Lin and Yiwei Dai and Yuxin Wang and Wenhao Chai and Shang Zhou and Dariush Wahdany and Ziyu She and Jiaming Hu and Zhikang Dong and Yuxuan Zhu and Sasha Cui and Ahson Saiyed and Arinbj{\"o}rn Kolbeinsson and Jesse Hu and Christopher Rytting and Ryan Marten and Yixin Wang and Alexandros G. Dimakis and A. Konwinski and Ludwig Schmidt},
 booktitle = {arXiv.org},
 journal = {ArXiv},
 title = {Terminal-Bench: Benchmarking Agents on Hard, Realistic Tasks in Command Line Interfaces},
 volume = {abs/2601.11868},
 year = {2026}
}

@Article{Li2024EEEBenchAC,
 author = {Ming Li and Jike Zhong and Tianle Chen and Yuxiang Lai and Konstantinos Psounis},
 booktitle = {Computer Vision and Pattern Recognition},
 journal = {2025 IEEE/CVF Conference on Computer Vision and Pattern Recognition (CVPR)},
 pages = {13337-13349},
 title = {EEE-Bench: A Comprehensive Multimodal Electrical And Electronics Engineering Benchmark},
 year = {2024}
}

@Article{Chen2025BenchmarkingLL,
 author = {Liangliang Chen and Zhihao Qin and Yiming Guo and Jacqueline Rohde and Ying Zhang},
 booktitle = {International Journal of Artificial Intelligence in Education},
 journal = {International Journal of Artificial Intelligence in Education},
 pages = {3294 - 3355},
 title = {Benchmarking Large Language Models on Homework Assessment in Circuit Analysis},
 volume = {35},
 year = {2025}
}

@article{Guo2025EngineeringAGI,
  title={Toward engineering AGI: Benchmarking the engineering design capabilities of LLMs},
  author={Guo, Xingang and Li, Yaxin and Kong, Xiangyi and Jiang, Yilan and Zhao, Xiayu and Gong, Zhihua and Zhang, Yufan and Li, Daixuan and Sang, Tianle and Zhu, Beixiao and others},
  journal={arXiv preprint arXiv:2509.16204},
  year={2025}
}

@inproceedings{Chaudhuri2024SPICED,
  title={Spiced: Syntactical bug and trojan pattern identification in a/ms circuits using llm-enhanced detection},
  author={Chaudhuri, Jayeeta and Thapar, Dhruv and Chaudhuri, Arjun and Firouzi, Farshad and Chakrabarty, Krishnendu},
  booktitle={2024 IEEE Physical Assurance and Inspection of Electronics (PAINE)},
  pages={1--7},
  year={2024},
  organization={IEEE}
}

@Article{Dolan2002BenchmarkingOA,
  author    = {Elizabeth D. Dolan and Jorge J. Mor{\'e}},
  title     = {Benchmarking optimization software with performance profiles},
  journal   = {Mathematical Programming},
  volume    = {91},
  number    = {2},
  pages     = {201--213},
  year      = {2002},
  publisher = {Springer}
}

@article{mialon2023gaia,
  title={GAIA: a benchmark for General AI Assistants},
  author={Mialon, Gr{\'e}goire and Fourrier, Cl{\'e}mentine and Swift, Craig and Wolf, Thomas and LeCun, Yann and Scialom, Thomas},
  journal={arXiv preprint arXiv:2311.12983},
  year={2023},
  url={https://arxiv.org/abs/2311.12983}
}

@article{Phan2025HumanitysLE,
  title={Humanity's Last Exam},
  author={Phan, Long and Gatti, Alice and Han, Ziwen and Li, Nathaniel and Hu, Josephina and Zhang, Hugh and Zhang, Chen Bo Calvin and Shaaban, Mohamed and Ling, John and Shi, Sean and others},
  journal={arXiv preprint arXiv:2501.14249},
  year={2025},
  url={https://arxiv.org/abs/2501.14249}
}

@misc{chowdhury2024swebenchverified,
  title={Introducing {SWE}-bench Verified},
  author={Chowdhury, Neil and Aung, James and Shern, Chan Jun and Jaffe, Oliver and Sherburn, Dane and Starace, Giulio and Mays, Evan and Dias, Rachel and Aljubeh, Marwan and Glaese, Mia and Jimenez, Carlos E. and Yang, John and Ho, Leyton and Patwardhan, Tejal and Liu, Kevin and Madry, Aleksander},
  year={2024},
  url={https://openai.com/index/introducing-swe-bench-verified/},
}

@article{novikov2025alphaevolve,
  title={AlphaEvolve: A coding agent for scientific and algorithmic discovery},
  author={Novikov, Alexander and V{\~u}, Ng{\^a}n and Eisenberger, Marvin and Dupont, Emilien and Huang, Po-Sen and Wagner, Adam Zsolt and Shirobokov, Sergey and Kozlovskii, Borislav and Ruiz, Francisco J. R. and Mehrabian, Abbas and Kumar, M. Pawan and See, Abigail and Chaudhuri, Swarat and Holland, George and Davies, Alex and Nowozin, Sebastian and Kohli, Pushmeet and Balog, Matej},
  journal={arXiv preprint arXiv:2506.13131},
  year={2025},
  url={https://arxiv.org/abs/2506.13131}
}

@article{yuksekgonul2026learning,
  title={Learning to Discover at Test Time},
  author={Yuksekgonul, Mert and Koceja, Daniel and Li, Xinhao and Bianchi, Federico and McCaleb, Jed and Wang, Xiaolong and Kautz, Jan and Choi, Yejin and Zou, James and Guestrin, Carlos and Sun, Yu},
  journal={arXiv preprint arXiv:2601.16175},
  year={2026},
  url={https://arxiv.org/abs/2601.16175}
}

@article{liu2025vision,
  title={A Vision for Auto Research with LLM Agents},
  author={Liu, Chengwei and Wang, Chong and Cao, Jiayue and Ge, Jingquan and Wang, Kun and Zhang, Lyuye and Cheng, Ming-Ming and Zhao, Penghai and Li, Tianlin and Jia, Xiaojun and Li, Xiang and Li, Xingshuai and Liu, Yang and Feng, Yebo and Huang, Yihao and Xu, Yijia and Sun, Yuqiang and Zhou, Zhenhong and Xu, Zhengzi},
  journal={arXiv preprint arXiv:2504.18765},
  year={2025},
  url={https://arxiv.org/abs/2504.18765}
}

@article{gao2025survey,
  title={A survey of self-evolving agents: On path to artificial super intelligence},
  author={Gao, Huan-ang and Geng, Jiayi and Hua, Wenyue and Hu, Mengkang and Juan, Xinzhe and Liu, Hongzhang and Liu, Shilong and Qiu, Jiahao and Qi, Xuan and Wu, Yiran and others},
  journal={arXiv preprint arXiv:2507.21046},
  volume={1},
  year={2025}
}

@inproceedings{liu2024agentbench,
  title={AgentBench: Evaluating LLMs as Agents},
  author={Liu, Xiao and Yu, Hao and Zhang, Hanchen and Xu, Yifan and Lei, Xuanyu and Lai, Hanyu and Gu, Yu and Ding, Hangliang and Men, Kaiwen and Yang, Kejuan and Zhang, Shudan and Deng, Xiang and Zeng, Aohan and Du, Zhengxiao and Zhang, Chenhui and Shen, Sheng and Zhang, Tianjun and Su, Yu and Sun, Huan and Huang, Minlie and Dong, Yuxiao and Tang, Jie},
  booktitle={The Twelfth International Conference on Learning Representations},
  year={2024}
}

@inproceedings{zhou2024webarena,
  title={WebArena: A Realistic Web Environment for Building Autonomous Agents},
  author={Zhou, Shuyan and Xu, Frank F. and Zhu, Hao and Zhou, Xuhui and Lo, Robert and Sridhar, Abishek and Cheng, Xianyi and Ou, Tianyue and Bisk, Yonatan and Fried, Daniel and Alon, Uri and Neubig, Graham},
  booktitle={The Twelfth International Conference on Learning Representations},
  year={2024}
}

@inproceedings{xie2024osworld,
  title={OSWorld: Benchmarking Multimodal Agents for Open-Ended Tasks in Real Computer Environments},
  author={Xie, Tianbao and Zhang, Danyang and Chen, Jixuan and Li, Xiaochuan and Zhao, Siheng and Cao, Ruisheng and Hua, Toh Jing and Cheng, Zhoujun and Shin, Dongchan and Lei, Fangyu and Liu, Yitao and Xu, Yiheng and Zhou, Shuyan and Savarese, Silvio and Xiong, Caiming and Zhong, Victor and Yu, Tao},
  booktitle={Advances in Neural Information Processing Systems},
  year={2024}
}

@inproceedings{yao2023react,
  title={ReAct: Synergizing Reasoning and Acting in Language Models},
  author={Yao, Shunyu and Zhao, Jeffrey and Yu, Dian and Du, Nan and Shafran, Izhak and Narasimhan, Karthik R. and Cao, Yuan},
  booktitle={The Eleventh International Conference on Learning Representations},
  year={2023}
}

@inproceedings{shinn2023reflexion,
  title={Reflexion: Language Agents with Verbal Reinforcement Learning},
  author={Shinn, Noah and Cassano, Federico and Gopinath, Ashwin and Narasimhan, Karthik and Yao, Shunyu},
  booktitle={Advances in Neural Information Processing Systems},
  year={2023}
}

@inproceedings{yao2023tree,
  title={Tree of Thoughts: Deliberate Problem Solving with Large Language Models},
  author={Yao, Shunyu and Yu, Dian and Zhao, Jeffrey and Shafran, Izhak and Griffiths, Thomas L. and Cao, Yuan and Narasimhan, Karthik},
  booktitle={Advances in Neural Information Processing Systems},
  year={2023}
}

@inproceedings{zhou2024lats,
  title={Language Agent Tree Search Unifies Reasoning, Acting, and Planning in Language Models},
  author={Zhou, Andy and Yan, Kai and Shlapentokh-Rothman, Michal and Wang, Haohan and Wang, Yu-Xiong},
  booktitle={Proceedings of the 41st International Conference on Machine Learning},
  pages={62138--62160},
  year={2024}
}

% ── Appendix (attention visualizations) ──────────────────────────────────────

\clearpage
\newpage
\appendix
\section{Detailed Task Catalog}
\label{app:task_catalog}

This section provides a detailed summary of the $47$ tasks in the Frontier-Eng benchmark, organized into five engineering categories. For each task, we specify the core objective, the scoring mechanism, and the safeguards implemented to ensure evaluation integrity.

\subsection{Computing \& Quantum Information (10 tasks)}

\paragraph{FlashAttention}
\begin{itemize}[leftmargin=2.2cm, labelsep=0.3cm, nosep]
    \item[\textbf{Objective.}] Implement a multi-head causal scaled dot-product attention forward kernel in CUDA or Triton.
    \item[\textbf{Scoring.}] The score is defined as $10^9$ divided by the geometric mean of kernel latency (ns) across multiple configurations, gated by a correctness check against PyTorch.
    \item[\textbf{Anti-hack.}] Correctness is verified on GPU tensors before timing; the evaluator and reference implementations are read-only.
\end{itemize}

\paragraph{MLA (Multi-Head Latent Attention)}
\begin{itemize}[leftmargin=2.2cm, labelsep=0.3cm, nosep]
    \item[\textbf{Objective.}] Implement a performance-optimized kernel for the MLA-style attention workload.
    \item[\textbf{Scoring.}] Score is $10^9$ divided by the geometric mean kernel latency (ns), gated by a correctness check against a reference implementation.
    \item[\textbf{Anti-hack.}] Evaluation runs in a sandboxed environment with independent correctness verification.
\end{itemize}

\paragraph{TriMul (Triangular Multiplicative Update)}
\begin{itemize}[leftmargin=2.2cm, labelsep=0.3cm, nosep]
    \item[\textbf{Objective.}] Implement a kernel for triangular multiplicative module operations involving masked batched tensor computations on GPU.
    \item[\textbf{Scoring.}] Latency-based performance score with support for hidden-seed leaderboard runs.
    \item[\textbf{Anti-hack.}] Correctness checks are executed in isolated worker processes to prevent interference.
\end{itemize}

\paragraph{MallocLab}
\begin{itemize}[leftmargin=2.2cm, labelsep=0.3cm, nosep]
    \item[\textbf{Objective.}] Implement a high-performance C dynamic memory allocator supporting \texttt{malloc}, \texttt{free}, and \texttt{realloc}.
    \item[\textbf{Scoring.}] A weighted blend of memory utilization and throughput (0--100) across allocation trace files.
    \item[\textbf{Anti-hack.}] The driver enforces heap invariants (alignment, bounds) and caps throughput at the libc reference level.
\end{itemize}

\paragraph{AES-128}
\begin{itemize}[leftmargin=2.2cm, labelsep=0.3cm, nosep]
    \item[\textbf{Objective.}] Implement AES-128 in CTR mode using C++.
    \item[\textbf{Scoring.}] Geometric mean throughput (Mbps) across random data streams, requiring an OpenSSL correctness gate.
    \item[\textbf{Anti-hack.}] Test vectors use random keys and plaintexts generated at runtime to prevent hard-coding.
\end{itemize}

\paragraph{SHA-256}
\begin{itemize}[leftmargin=2.2cm, labelsep=0.3cm, nosep]
    \item[\textbf{Objective.}] Implement the SHA-256 hashing algorithm in C++.
    \item[\textbf{Scoring.}] Throughput-based score (Mbps) verified against the OpenSSL reference implementation.
    \item[\textbf{Anti-hack.}] Uses random-length inputs and frozen verification code in a temporary sandbox.
\end{itemize}

\paragraph{SHA3-256}
\begin{itemize}[leftmargin=2.2cm, labelsep=0.3cm, nosep]
    \item[\textbf{Objective.}] Implement the SHA3-256 (Keccak) hashing algorithm in C++.
    \item[\textbf{Scoring.}] Throughput (Mbps) with correctness validated via OpenSSL EVP interfaces.
    \item[\textbf{Anti-hack.}] The evaluator parses detailed pass/fail counts to detect partial correctness edge cases.
\end{itemize}

\paragraph{Routing QFTEntangled}
\begin{itemize}[leftmargin=2.2cm, labelsep=0.3cm, nosep]
    \item[\textbf{Objective.}] Optimize quantum circuit routing for QFT-entangled circuits on the IBM Falcon 27-qubit topology.
    \item[\textbf{Scoring.}] Normalized cost (0--3) based on two-qubit gate count and circuit depth after canonical transpilation.
    \item[\textbf{Anti-hack.}] Uses evaluator-owned transpilation scripts and MQT Bench reference circuits.
\end{itemize}

\paragraph{Clifford+T Synthesis}
\begin{itemize}[leftmargin=2.2cm, labelsep=0.3cm, nosep]
    \item[\textbf{Objective.}] Synthesize QFT circuits into the Clifford+T gate set while minimizing structural costs.
    \item[\textbf{Scoring.}] A cost function minimizing the count of T-gates, two-qubit gates, and overall circuit depth.
    \item[\textbf{Anti-hack.}] Scoring is performed against MQT Bench optimization anchors; verifier code is read-only.
\end{itemize}

\paragraph{Cross-Target QAOA}
\begin{itemize}[leftmargin=2.2cm, labelsep=0.3cm, nosep]
    \item[\textbf{Objective.}] Optimize QAOA circuits for simultaneous evaluation on IBM Falcon and IonQ Aria hardware backends.
    \item[\textbf{Scoring.}] Mean normalized cost across hardware targets to ensure cross-backend robustness.
    \item[\textbf{Anti-hack.}] Multi-target evaluation prevents overfitting to a single device's topology.
\end{itemize}

\subsection{Operations Research \& Decision Science (9 tasks)}

\paragraph{tree\_gsm\_safety\_stock}
\begin{itemize}[leftmargin=2.2cm, labelsep=0.3cm, nosep]
    \item[\textbf{Objective.}] Optimize committed service times (CST) in tree-structured multi-echelon supply networks to minimize holding costs.
    \item[\textbf{Scoring.}] Weighted composite of cost improvement, robustness under stress, SLA compliance, and solution simplicity.
    \item[\textbf{Anti-hack.}] Evaluator uses an independent network model and the \texttt{stockpyl} library for cost computation.
\end{itemize}

\paragraph{general\_meio}
\begin{itemize}[leftmargin=2.2cm, labelsep=0.3cm, nosep]
    \item[\textbf{Objective.}] Set base-stock levels for a 5-node directed supply network under stochastic Poisson demand.
    \item[\textbf{Scoring.}] Blend of cost efficiency, service level, and robustness evaluated via Monte Carlo simulation.
    \item[\textbf{Anti-hack.}] Fixed network topology, cost parameters, and simulation seeds in a read-only evaluator.
\end{itemize}

\paragraph{joint\_replenishment}
\begin{itemize}[leftmargin=2.2cm, labelsep=0.3cm, nosep]
    \item[\textbf{Objective.}] Determine base cycle time and integer order multiples for 8 SKUs sharing a fixed setup cost.
    \item[\textbf{Scoring.}] Cost improvement relative to an independent EOQ baseline and coordination rewards.
    \item[\textbf{Anti-hack.}] Analytical cost formulas and problem parameters are hardcoded in the read-only evaluator.
\end{itemize}

\paragraph{finite\_horizon\_dp}
\begin{itemize}[leftmargin=2.2cm, labelsep=0.3cm, nosep]
    \item[\textbf{Objective.}] Design a time-varying (s, S) ordering policy for an 8-period stochastic inventory problem.
    \item[\textbf{Scoring.}] Monte Carlo evaluation of cost improvement, fill rate targets, and order cadence.
    \item[\textbf{Anti-hack.}] Uses fixed simulation parameters and a dynamic programming reference solution.
\end{itemize}

\paragraph{disruption\_eoqd}
\begin{itemize}[leftmargin=2.2cm, labelsep=0.3cm, nosep]
    \item[\textbf{Objective.}] Determine the optimal order quantity Q under supply disruption conditions.
    \item[\textbf{Scoring.}] Model cost vs classic EOQ baseline and simulated fill rate performance.
    \item[\textbf{Anti-hack.}] Fixed cost models and simulation parameters in a read-only verification environment.
\end{itemize}

\paragraph{abz}
\begin{itemize}[leftmargin=2.2cm, labelsep=0.3cm, nosep]
    \item[\textbf{Objective.}] Minimize makespan on classical ABZ family Job-Shop Scheduling Problem (JSSP) instances.
    \item[\textbf{Scoring.}] Score is defined as $\min(100, 100 \times \text{target} / \text{makespan})$ where target is the known optimum.
    \item[\textbf{Anti-hack.}] Schedules are fully validated for machine, precedence, and duration constraints.
\end{itemize}

\paragraph{swv}
\begin{itemize}[leftmargin=2.2cm, labelsep=0.3cm, nosep]
    \item[\textbf{Objective.}] Minimize makespan on SWV family JSSP instances (up to 50$\times$10).
    \item[\textbf{Scoring.}] Identical ratio-based scoring formula and validation mechanism as the ABZ task.
    \item[\textbf{Anti-hack.}] Instance data and the evaluator are marked read-only to prevent manipulation.
\end{itemize}

\paragraph{ta}
\begin{itemize}[leftmargin=2.2cm, labelsep=0.3cm, nosep]
    \item[\textbf{Objective.}] Minimize makespan on large-scale Taillard JSSP instances (up to 100$\times$20).
    \item[\textbf{Scoring.}] Mean score across all instances in the family; only pure Python solutions are allowed.
    \item[\textbf{Anti-hack.}] Makespan is recomputed from the submitted schedule; no external solvers permitted.
\end{itemize}

\paragraph{robust\_mvo\_rebalance}
\begin{itemize}[leftmargin=2.2cm, labelsep=0.3cm, nosep]
    \item[\textbf{Objective.}] Solve a single-period robust mean-variance portfolio rebalancing problem under multiple constraints.
    \item[\textbf{Scoring.}] Normalized objective value vs CVXPY optimum, with heavy penalties for feasibility violations.
    \item[\textbf{Anti-hack.}] Reference optimum is recomputed per instance; strict independent constraint checking.
\end{itemize}

\subsection{Robotics, Control \& Energy Systems (8 tasks)}

\paragraph{DynamicObstacleAvoidanceNavigation}
\begin{itemize}[leftmargin=2.2cm, labelsep=0.3cm, nosep]
    \item[\textbf{Objective.}] Plan open-loop velocity commands for a differential-drive robot navigating static and dynamic obstacles.
    \item[\textbf{Scoring.}] Success-gated inverse arrival time across three fixed 2D scenes.
    \item[\textbf{Anti-hack.}] Re-simulation using a unicycle model against fixed data; acceleration limits enforced.
\end{itemize}

\paragraph{PIDTuning}
\begin{itemize}[leftmargin=2.2cm, labelsep=0.3cm, nosep]
    \item[\textbf{Objective.}] Tune 12 PID gains for a 2D quadrotor across multiple flight scenarios with wind disturbances.
    \item[\textbf{Scoring.}] Geometric mean of inverse ITAE cost; any infeasible scenario zeros the total score.
    \item[\textbf{Anti-hack.}] Full flight dynamics (motor lag, drag, torque limits) run in a read-only evaluator.
\end{itemize}

\paragraph{QuadrupedGaitOptimization}
\begin{itemize}[leftmargin=2.2cm, labelsep=0.3cm, nosep]
    \item[\textbf{Objective.}] Optimize 8 gait parameters for a MuJoCo Ant model to maximize forward speed.
    \item[\textbf{Scoring.}] Average forward speed (m/s) subject to roll/pitch stability and actuator force limits.
    \item[\textbf{Anti-hack.}] Measured directly from physics rollout; model XML and gait configurations are read-only.
\end{itemize}

\paragraph{RobotArmCycleTimeOptimization}
\begin{itemize}[leftmargin=2.2cm, labelsep=0.3cm, nosep]
    \item[\textbf{Objective.}] Minimize motion time for a 7-DOF KUKA arm avoiding a box obstacle.
    \item[\textbf{Scoring.}] Inverse cycle time verified via cubic spline fitting and PyBullet collision queries.
    \item[\textbf{Anti-hack.}] Dense sampling check for joint limits and velocity/acceleration constraints.
\end{itemize}

\paragraph{UAVInspectionCoverageWithWind}
\begin{itemize}[leftmargin=2.2cm, labelsep=0.3cm, nosep]
    \item[\textbf{Objective.}] Fly a UAV to cover inspection points while avoiding no-fly zones under wind disturbances.
    \item[\textbf{Scoring.}] Weighted blend of coverage ratio and energy consumption (acceleration integral).
    \item[\textbf{Anti-hack.}] Coverage and energy recomputed from acceleration commands via physics model.
\end{itemize}

\paragraph{BatteryFastChargingProfile}
\begin{itemize}[leftmargin=2.2cm, labelsep=0.3cm, nosep]
    \item[\textbf{Objective.}] Design a multi-stage constant-current charging profile for a lithium-ion cell.
    \item[\textbf{Scoring.}] Weighted combination of charge time, degradation, peak temperature, and voltage softness.
    \item[\textbf{Anti-hack.}] Simulator uses frozen physics parameters; profile bounds validated before execution.
\end{itemize}

\paragraph{BatteryFastChargingSPMe}
\begin{itemize}[leftmargin=2.2cm, labelsep=0.3cm, nosep]
    \item[\textbf{Objective.}] Optimize charging under a high-fidelity SPMe electrochemical-thermal model.
    \item[\textbf{Scoring.}] Weighted combination of time, aging, plating, and thermal scores (0--100).
    \item[\textbf{Anti-hack.}] Candidate supplies only the policy; all physics logic resides in the read-only evaluator.
\end{itemize}

\paragraph{hand\_written\_control}
\begin{itemize}[leftmargin=2.2cm, labelsep=0.3cm, nosep]
    \item[\textbf{Objective.}] Write a joint load-shifting, cooling, and battery control policy for a sustainable data center.
    \item[\textbf{Scoring.}] Average across four scenarios of $100\sqrt{0.85\,\Delta_\text{carbon} + 0.15\,\Delta_\text{water}}$ with linear penalties for dropped/overdue tasks.
    \item[\textbf{Anti-hack.}] Actions validated against discrete space; improvement measured on identical seeds.
\end{itemize}

\subsection{Optics \& Communication Systems (10 tasks)}

\paragraph{adaptive\_fault\_tolerant\_fusion}
\begin{itemize}[leftmargin=2.2cm, labelsep=0.3cm, nosep]
    \item[\textbf{Objective.}] Fuse corrupted wavefront sensor slopes for multi-sensor adaptive optics control.
    \item[\textbf{Scoring.}] Weighted utility of mean RMS error and Strehl ratio across 320 turbulence scenarios.
    \item[\textbf{Anti-hack.}] True phase and clean slopes hidden; commands checked for shape and voltage limits.
\end{itemize}

\paragraph{adaptive\_temporal\_smooth\_control}
\begin{itemize}[leftmargin=2.2cm, labelsep=0.3cm, nosep]
    \item[\textbf{Objective.}] Sequential AO control balancing correction quality against command smoothness.
    \item[\textbf{Scoring.}] Utility function emphasizing temporal smoothness (mean slew) and Strehl ratio.
    \item[\textbf{Anti-hack.}] Hidden DM surface model; only delayed/noisy slopes are available to the agent.
\end{itemize}

\paragraph{phase\_dammann\_uniform\_orders}
\begin{itemize}[leftmargin=2.2cm, labelsep=0.3cm, nosep]
    \item[\textbf{Objective.}] Optimize binary phase transition positions for uniform Dammann grating diffraction.
    \item[\textbf{Scoring.}] Composite score of uniformity, efficiency, and balance across orders $-3$ to $+3$.
    \item[\textbf{Anti-hack.}] Intensities computed via \texttt{diffractio} simulation; literature-based oracle comparison.
\end{itemize}

\paragraph{phase\_fourier\_pattern\_holography}
\begin{itemize}[leftmargin=2.2cm, labelsep=0.3cm, nosep]
    \item[\textbf{Objective.}] Design a 2D phase pattern for a Fourier hologram with high energy and dark-zone suppression.
    \item[\textbf{Scoring.}] Pattern match fidelity, target zone energy, and dark-zone suppression ratio.
    \item[\textbf{Anti-hack.}] Independent forward model intensity computation; read-only oracle comparison.
\end{itemize}

\paragraph{fiber\_wdm\_channel\_power\_allocation}
\begin{itemize}[leftmargin=2.2cm, labelsep=0.3cm, nosep]
    \item[\textbf{Objective.}] Assign users to WDM channels and set launch power to optimize throughput.
    \item[\textbf{Scoring.}] Satisfaction, BER pass rate, spectral utilization, and SNR-based terms.
    \item[\textbf{Anti-hack.}] Strict assignment rules; SNR and BER recomputed from interference model.
\end{itemize}

\paragraph{fiber\_mcs\_power\_scheduling}
\begin{itemize}[leftmargin=2.2cm, labelsep=0.3cm, nosep]
    \item[\textbf{Objective.}] Jointly select MCS and per-user transmit power under a global linear power budget.
    \item[\textbf{Scoring.}] Weighted satisfaction, BER pass ratio, and spectral efficiency.
    \item[\textbf{Anti-hack.}] Power and MCS bounds enforced; all metrics recomputed by the evaluator.
\end{itemize}

\paragraph{fiber\_guardband\_spectrum\_packing}
\begin{itemize}[leftmargin=2.2cm, labelsep=0.3cm, nosep]
    \item[\textbf{Objective.}] Pack 24 users into spectrum slots with guard-bands while meeting BER requirements.
    \item[\textbf{Scoring.}] Acceptance ratio, utilization, compactness, and BER pass ratio.
    \item[\textbf{Anti-hack.}] Geometry (overlap, guard slots) and SNR independently validated.
\end{itemize}

\paragraph{holographic\_multifocus\_power\_ratio}
\begin{itemize}[leftmargin=2.2cm, labelsep=0.3cm, nosep]
    \item[\textbf{Objective.}] Design phase layers matching target power ratios at six focal spots in a single plane.
    \item[\textbf{Scoring.}] Efficiency score, ratio match, and shape cosine fidelity.
    \item[\textbf{Anti-hack.}] Metrics derived from simulated wave propagation using \texttt{torchoptics}.
\end{itemize}

\paragraph{holographic\_multiplane\_focusing}
\begin{itemize}[leftmargin=2.2cm, labelsep=0.3cm, nosep]
    \item[\textbf{Objective.}] Multi-plane holographic focusing with per-plane spot centers and target power ratios.
    \item[\textbf{Scoring.}] Average of per-plane efficiency, ratio, and shape cosine scores.
    \item[\textbf{Anti-hack.}] Evaluator propagates light fields to each plane independently.
\end{itemize}

\paragraph{HighReliableSimulation}
\begin{itemize}[leftmargin=2.2cm, labelsep=0.3cm, nosep]
    \item[\textbf{Objective.}] Implement a variance-controlled importance sampling BER estimator for Hamming codes.
    \item[\textbf{Scoring.}] Trade-off between estimation accuracy and runtime compared to a calibrated reference.
    \item[\textbf{Anti-hack.}] Decoder and code constructed by evaluator; RNG state reset post-construction.
\end{itemize}

\subsection{Physical Sciences \& Engineering Design (10 tasks)}

\paragraph{ISCSO2015}
\begin{itemize}[leftmargin=2.2cm, labelsep=0.3cm, nosep]
    \item[\textbf{Objective.}] Minimize structural weight of a 45-bar 2D truss under stress and displacement limits.
    \item[\textbf{Scoring.}] Negative structural weight verified by an independent 2D FEM solver.
    \item[\textbf{Anti-hack.}] Weight and constraint satisfaction come from evaluator's FEM; isolated temp directory.
\end{itemize}

\paragraph{ISCSO2023}
\begin{itemize}[leftmargin=2.2cm, labelsep=0.3cm, nosep]
    \item[\textbf{Objective.}] Minimize weight of a 284-member 3D tower with discrete section selection.
    \item[\textbf{Scoring.}] FEM-based weight score gated by a self-reported evaluation budget check.
    \item[\textbf{Anti-hack.}] Independent 3D FEM solve; strict handling of exit codes and stderr output.
\end{itemize}

\paragraph{TopologyOptimization}
\begin{itemize}[leftmargin=2.2cm, labelsep=0.3cm, nosep]
    \item[\textbf{Objective.}] Optimize element densities for a 2D MBB beam to minimize compliance.
    \item[\textbf{Scoring.}] Negative compliance subject to a strict volume fraction constraint ($\le 0.50$).
    \item[\textbf{Anti-hack.}] Built-in FEM solve; evaluator does not trust candidate-reported compliance values.
\end{itemize}

\paragraph{snar\_multiobjective}
\begin{itemize}[leftmargin=2.2cm, labelsep=0.3cm, nosep]
    \item[\textbf{Objective.}] Pareto-optimize SnAr reaction yield vs environmental factor across 3 seeds.
    \item[\textbf{Scoring.}] 2D hypervolume in a normalized minimization space (24 experiments per seed).
    \item[\textbf{Anti-hack.}] Objective values derived from Summit benchmark emulator; read-only scoring.
\end{itemize}

\paragraph{mit\_case1\_mixed}
\begin{itemize}[leftmargin=2.2cm, labelsep=0.3cm, nosep]
    \item[\textbf{Objective.}] Maximize reaction yield with mixed continuous and categorical variables (MIT kinetics).
    \item[\textbf{Scoring.}] Best achieved yield clamped to the [0, 1] interval.
    \item[\textbf{Anti-hack.}] Summit emulator provides true yields; task definition and scoring are read-only.
\end{itemize}

\paragraph{reizman\_suzuki\_pareto}
\begin{itemize}[leftmargin=2.2cm, labelsep=0.3cm, nosep]
    \item[\textbf{Objective.}] Pareto-optimize Suzuki coupling yield vs turnover number.
    \item[\textbf{Scoring.}] 2D hypervolume based on yield and turnover (normalized by 100 and 200).
    \item[\textbf{Anti-hack.}] Summit-backed evaluation with a fixed, read-only scoring rubric.
\end{itemize}

\paragraph{MannedLunarLanding}
\begin{itemize}[leftmargin=2.2cm, labelsep=0.3cm, nosep]
    \item[\textbf{Objective.}] Design a CRTBP Earth--Moon trajectory to maximize delivered lunar payload.
    \item[\textbf{Scoring.}] Payload mass gated by time monotonicity, fuel bookkeeping, and altitude constraints.
    \item[\textbf{Anti-hack.}] Numerical integration via Octave; physics-based validation of state vectors.
\end{itemize}

\paragraph{CarAerodynamicsSensing}
\begin{itemize}[leftmargin=2.2cm, labelsep=0.3cm, nosep]
    \item[\textbf{Objective.}] Select 30 sensor locations on a car mesh for surface pressure field reconstruction.
    \item[\textbf{Scoring.}] Reconstruction accuracy of a frozen Transolver model on held-out CFD cases.
    \item[\textbf{Anti-hack.}] Server-side inference; read-only model weights and surface mesh data.
\end{itemize}

\paragraph{predict\_modality}
\begin{itemize}[leftmargin=2.2cm, labelsep=0.3cm, nosep]
    \item[\textbf{Objective.}] Predict ADT modality from RNA measurements on the BMMC CITE dataset.
    \item[\textbf{Scoring.}] Mean of correlation score and error score (RMSE) against held-out ground truth.
    \item[\textbf{Anti-hack.}] Ground truth downloaded from fixed S3 URL; dataset consistency checks.
\end{itemize}

\paragraph{EngDesign}
\begin{itemize}[leftmargin=2.2cm, labelsep=0.3cm, nosep]
    \item[\textbf{Objective.}] Resolve 7 heterogeneous sub-problems (drivers, denoising, CPU logic, robotics, topology).
    \item[\textbf{Scoring.}] Unweighted arithmetic mean of seven independent sub-scores.
    \item[\textbf{Anti-hack.}] FEA re-solve for topology; re-simulation for robotics; point-by-point ISA verification.
\end{itemize}

\section{Case Study}

This appendix examines four representative tasks: \texttt{ComputerSystems\_MallocLab}, \texttt{EnergyStorage\_BatteryFastChargingProfile}, \texttt{InventoryOptimization\_general\_meio}, and \texttt{Cryptographic\_SHA3-256}. The objective is to complement aggregate statistics with trajectory-level evidence showing how model capability and search mechanism interact under different task structures. To preserve visual clarity, the figures use no-text running-best staircase plots: gray points denote all evaluated candidates, and the green staircase denotes the best score observed up to each experiment index.

\begin{table*}[t]
\centering
\small
\caption{Final-score comparison across models and frameworks for the four focus cases.}
\label{tab:evolve-focus-case-summary-en}
\begin{tabular}{llrrr}
\toprule
Task & Best Combination & Final Score & Runner-up Combination & Final Score \\
\midrule
\texttt{MallocLab} & \texttt{Claude + OpenEvolve} & 97.00 & \texttt{Claude + ShinkaEvolve} & 97.00 \\
\texttt{BatteryFastChargingProfile} & \texttt{Claude + OpenEvolve} & 120.80 & \texttt{Claude + ABMCTS} & 113.53 \\
\texttt{InventoryOptimization\_general\_meio} & \texttt{Claude + OpenEvolve} & 0.9929 & \texttt{Claude + ShinkaEvolve} & 0.9694 \\
\texttt{SHA3-256} & \texttt{Claude + OpenEvolve} & 28.90 & \texttt{GPT-OSS + ABMCTS} & 24.42 \\
\bottomrule
\end{tabular}
\end{table*}

\subsection{\texttt{ComputerSystems\_MallocLab}}

\paragraph{Task characteristics.}
\texttt{MallocLab} requires crossing a structural threshold: moving from a naive bump allocator to a production-grade implementation with free-block reuse, coalescing, and metadata management.

\paragraph{Empirical pattern.}
We observe a stark \textbf{granularity-capability gap}. Claude is indifferent to search dynamics, achieving a score of 97 via both sudden architectural jumps (\texttt{openevolve}) and incremental refinement (\texttt{shinkaevolve}). In contrast, GPT-OSS suffers a "capability collapse" under large-scale rewrites (stopping at 53), but recovers to 86 when the same structural transition is decomposed into fine-grained steps by \texttt{shinkaevolve}.

\paragraph{Logged code changes.}
The logs reveal two distinct evolutionary paths. Claude performs a "one-shot" architectural overhaul, replacing the entire codebase with a complex segregated-list design in a single step. GPT-OSS fails this leap, producing buggy or simplistic implementations when forced to rewrite. However, under \texttt{shinkaevolve}, GPT-OSS successfully navigates the transition through a sequence of atomic refinements—moving from headers to coalescing, then to segregated classes.

\paragraph{Implication.}
In systems engineering, the framework acts as a \textbf{reasoning stabilizer}. Frontier models possess the "zero-shot" density to leapfrog architectural thresholds directly. For mid-tier models, the framework's granularity is the difference between architectural collapse and successful evolution; without incremental scaffolding, they remain trapped behind structural barriers that their raw reasoning cannot bridge.

\begin{figure*}[t]
\centering
\includegraphics[width=0.94\textwidth]{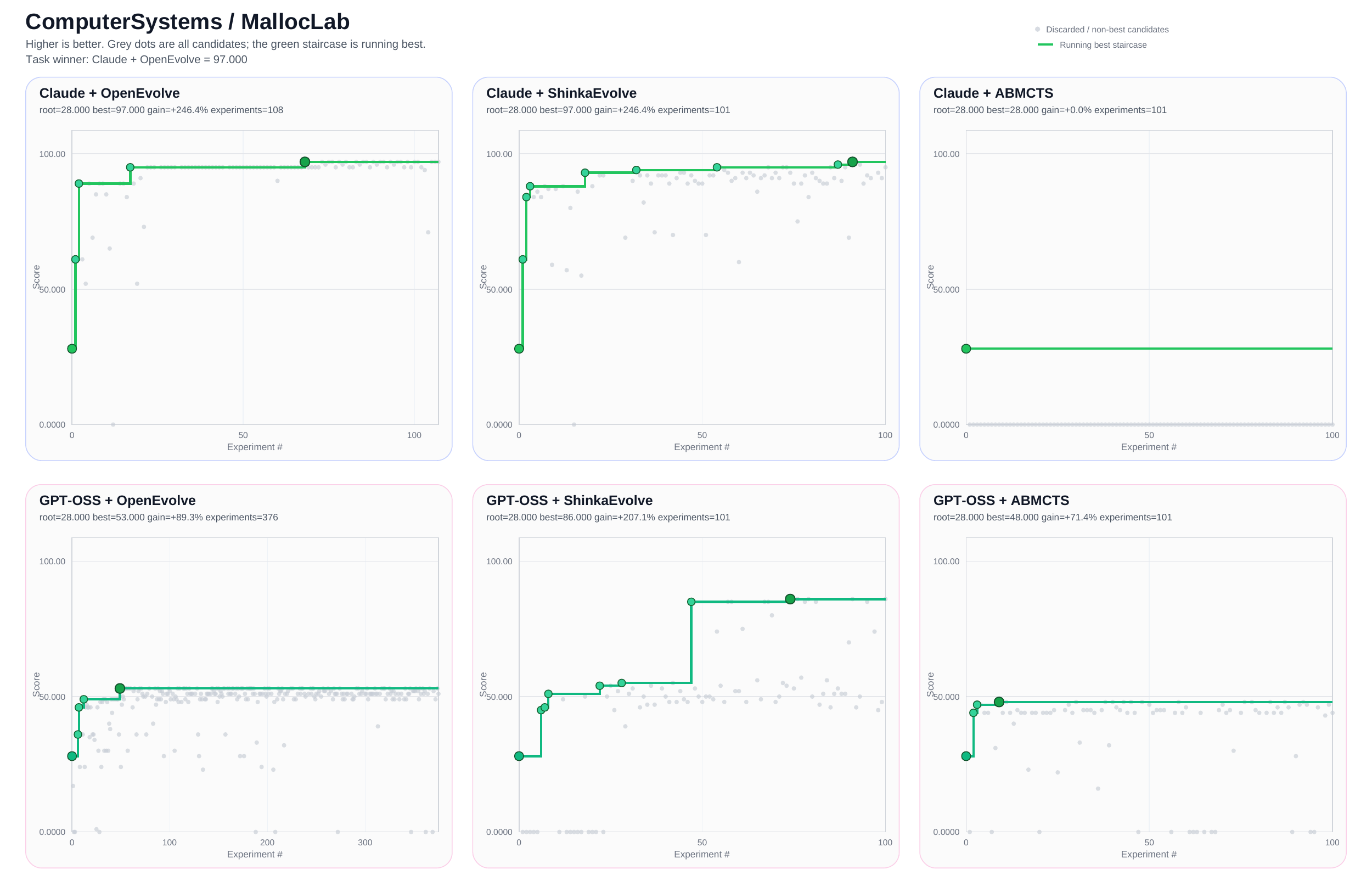}
% TODO: Missing figure file: figures/ComputerSystems_MallocLab__step_plot_notext.pdf (or .pdf)
\caption{Running-best staircase plots for \texttt{MallocLab}.}
\label{fig:evolve-focus-malloclab-en}
\end{figure*}

\subsection{\texttt{EnergyStorage\_BatteryFastChargingProfile}}

\paragraph{Task characteristics.}
This is a boundary-sensitive control problem. The challenge is not structural redesign but aggressive exploitation of safety headroom (voltage and temperature limits) without triggering hard constraints.

\paragraph{Empirical pattern.}
This task favors the "boldness" of \texttt{openevolve}. Claude + \texttt{openevolve} (120.80) enters the near-optimal region almost immediately, while incremental methods (\texttt{shinkaevolve}) remain trapped in conservative regimes, reaching only 87.58.

\paragraph{Logged code changes.}
Claude + \texttt{openevolve} immediately abandons the safe four-stage baseline for an aggressive six-stage profile, explicitly citing "low-SOC voltage headroom" as the justification for higher currents. In contrast, incremental runs (\texttt{shinkaevolve}) spend their entire budget making minor adjustments to the switch points of the conservative baseline, never gaining the "courage" to shift the current levels upward.

\paragraph{Implication.}
Optimization is not a search for safety, but a \textbf{calculated invasion of the safety margin}. \texttt{openevolve}'s success demonstrates that when models are freed from the inertia of local refinement, they can immediately identify and occupy the near-optimal boundary—a feat that incremental methods fail to replicate due to their inherent conservatism.

\begin{figure*}[t]
\centering
\includegraphics[width=0.94\textwidth]{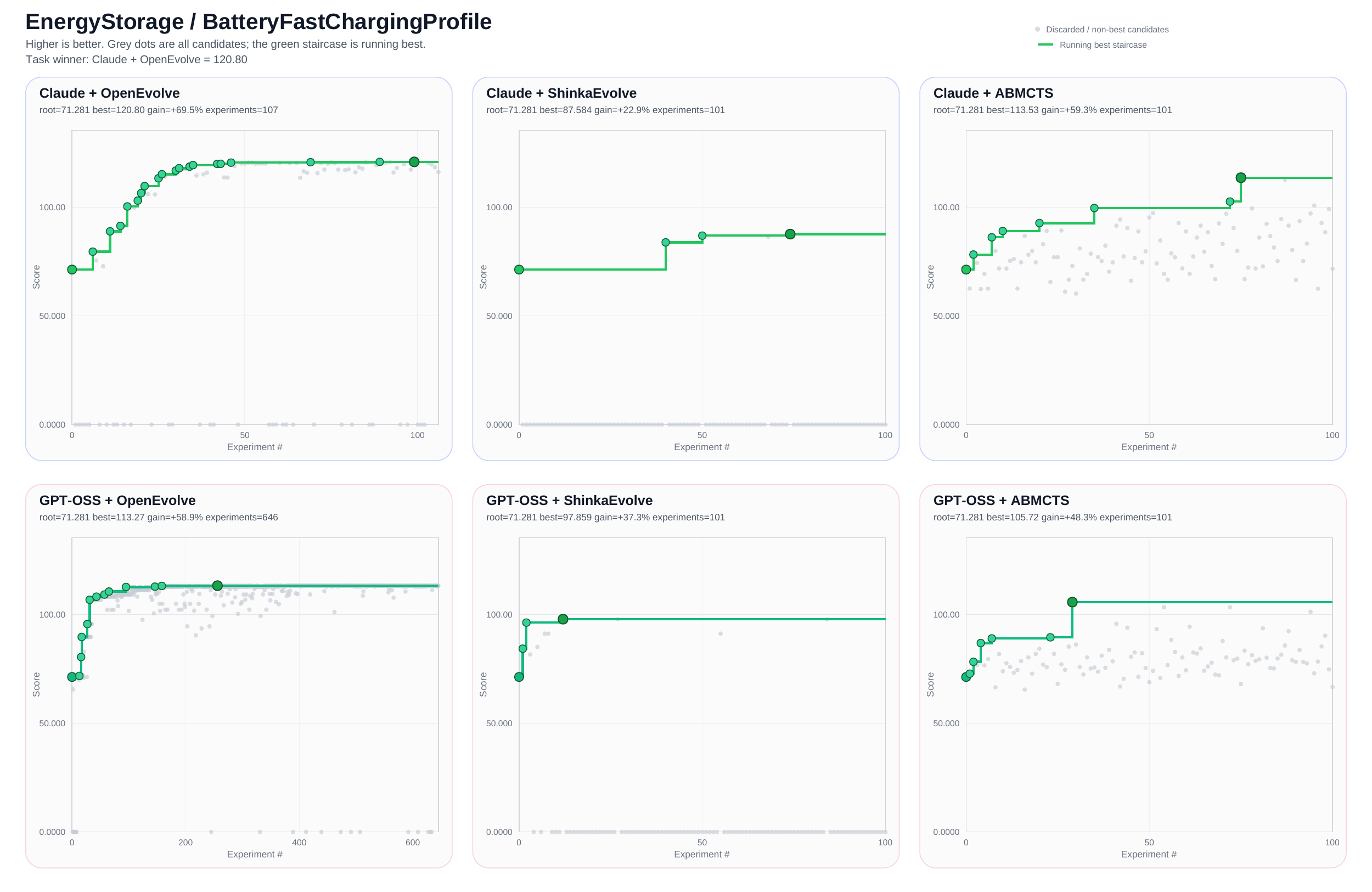}
\caption{Running-best staircase plots for \texttt{BatteryFastChargingProfile}.}
\label{fig:evolve-focus-battery-en}
\end{figure*}

\subsection{\texttt{InventoryOptimization\_general\_meio}}

\paragraph{Task characteristics.}
A multi-echelon inventory problem where local heuristics fail due to coupled supply-chain dynamics. Success requires discovering a unified network-level policy.

\paragraph{Empirical pattern.}
Claude + \texttt{openevolve} reaches 0.9929 by replacing the entire logic early. While refinement helps (Claude + \texttt{shinkaevolve} reaches 0.9694), the largest gains come from the initial policy-structure shift rather than continuous micro-tuning.

\paragraph{Logged code changes.}
The strongest trajectories do not merely perturb stock levels; they perform a "regime shift." Claude + \texttt{openevolve} replaces the baseline with a custom two-phase procedure using explicit local holding costs and node-specific stockout penalties. Incremental runs eventually reach similar logic, but only after dozens of steps spent struggling with the limitations of the original heuristic.

\paragraph{Implication.}
Engineering complex systems is less a tuning exercise and more a \textbf{policy-discovery problem}. The dominance of \texttt{openevolve} suggests that the most effective way to solve coupled dynamics is to replace local heuristics with a unified, solver-aligned policy structure—a "regime shift" that incremental methods often fail to trigger in time.

\begin{figure*}[t]
\centering
\includegraphics[width=0.94\textwidth]{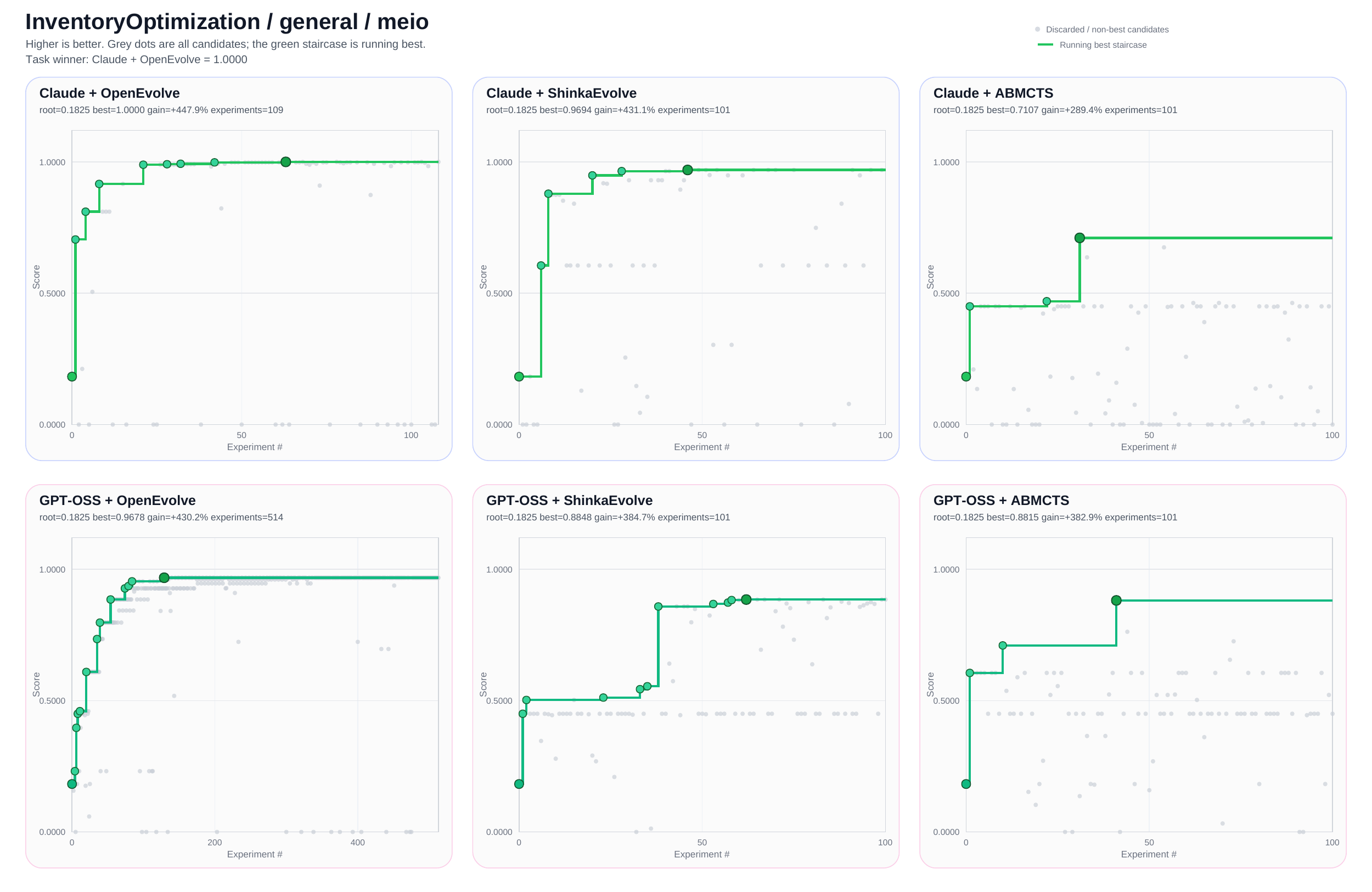}
\caption{Running-best staircase plots for \texttt{InventoryOptimization\_general\_meio}.}
\label{fig:evolve-focus-meio-en}
\end{figure*}

\subsection{\texttt{Cryptographic\_SHA3-256}}

\paragraph{Task characteristics.}
A pure implementation-level performance task. Since the algorithm is fixed, the only variable is the density and efficiency of the C++ implementation.

\paragraph{Empirical pattern.}
This task still favors large implementation-level rewrites, but the gap is narrower than previously stated. Claude + \texttt{openevolve} reaches 28.90, ahead of GPT-OSS + \texttt{abmcts} at 24.42 and Claude + \texttt{shinkaevolve} at 22.12.

\paragraph{Logged code changes.}
The logs show that the winning rewrite is a total reconstruction: moving from stream-based I/O to \texttt{mmap}-based file handling and replacing high-level abstractions with tightly scalarized, 64-bit word-level Keccak rounds. GPT-OSS and incremental runs attempt to optimize the existing structure (e.g., adding buffers), but these "patches" cannot compete with the raw throughput of a perfectly engineered kernel.

\paragraph{Implication.}
Low-level optimization remains a \textbf{high-sensitivity game of implementation density}. The leading score comes from a stronger end-to-end rewrite, but the local results suggest a competitive band rather than a massive separation: multiple combinations improve on the baseline, yet only the strongest rewrites break clearly into the top tier.

\begin{figure*}[t]
\centering
\includegraphics[width=0.94\textwidth]{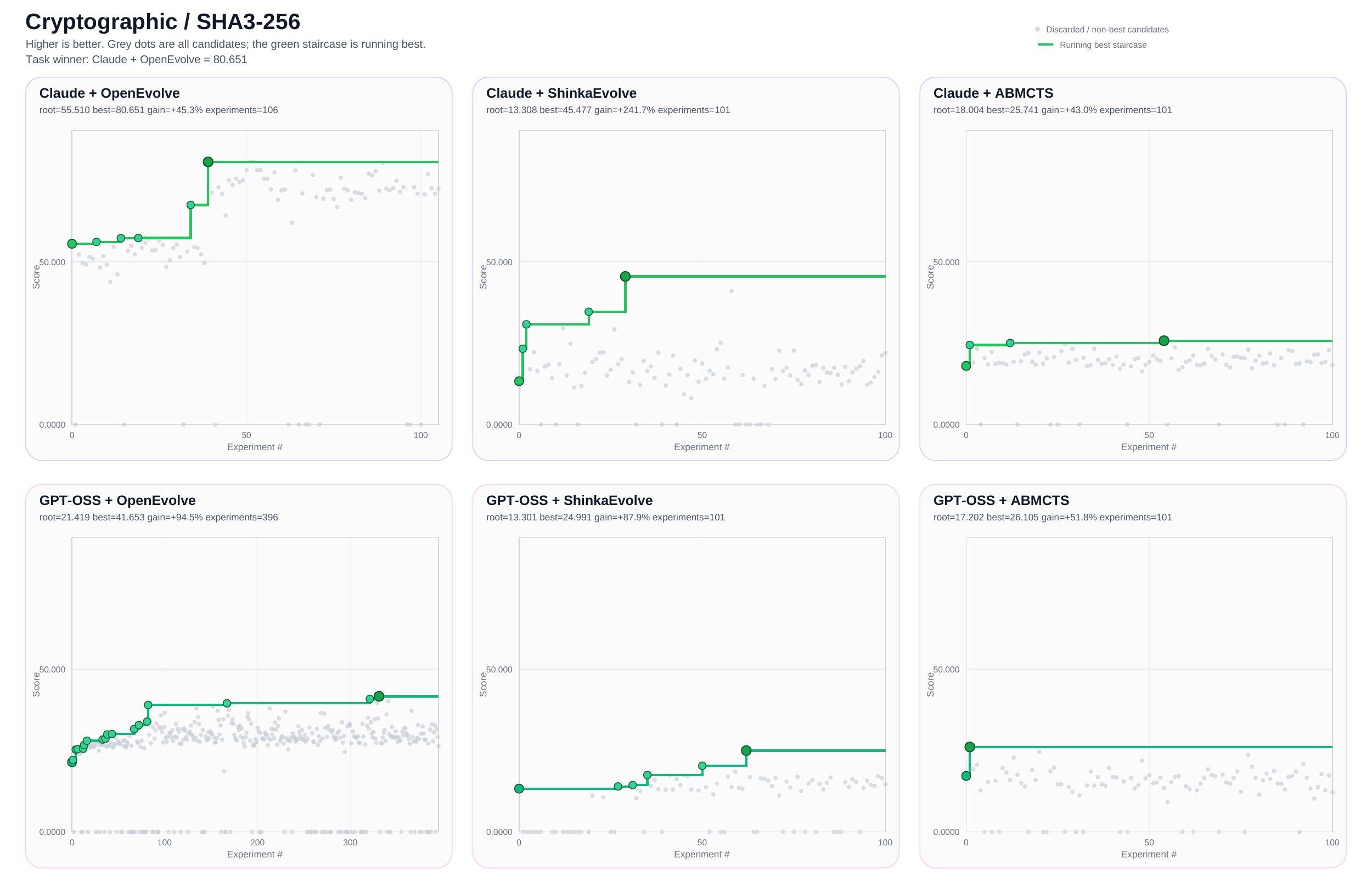}
\caption{Running-best staircase plots for \texttt{SHA3-256}.}
\label{fig:evolve-focus-sha3-en}
\end{figure*}

\subsection{Cross-Case Synthesis}

Taken together, the four cases support three conclusions. First, task structure determines trajectory shape: \texttt{MallocLab} is governed by a structural threshold, \texttt{BatteryFastChargingProfile} by boundary-sensitive control, \texttt{general\_meio} by policy-structure search, and \texttt{SHA3-256} by implementation-level performance engineering. Second, \texttt{openevolve} is not strong for the same reason in every task, but a recurring property is its ability to surface high-value candidates early. Third, Claude is systematically stronger than GPT-OSS across all four tasks, with the largest margins appearing where success requires either substantial code rewriting or more accurate use of available design headroom.

The resulting picture is therefore not that one framework is uniformly best. Rather, model capability sets the ceiling of candidate quality, framework dynamics determine how search approaches that ceiling, and their interaction depends on whether the underlying task is structural, control-oriented, policy-oriented, or implementation-level.

\section{Raw Data for Model Comparison}
\label{app:raw_pivot_data}

For completeness, Table~\ref{tab:raw-pivot-data-en} reports the raw per-task values underlying the model comparison in Section~\ref{sec:openevolve-model-comparison}, including the rank results in Table~\ref{tab:openevolve-models-rank-en}.

\begin{table*}[p]
\centering
\caption{Raw per-task scores underlying the Experiment~1 model comparison in Section~\ref{sec:openevolve-model-comparison} (Table~\ref{tab:openevolve-models-rank-en}).}
\label{tab:raw-pivot-data-en}
\small
\setlength{\tabcolsep}{3pt}
\renewcommand{\arraystretch}{1.06}
\resizebox{\textwidth}{!}{%
\begin{tabular}{@{}lrrrrrrrrrr@{}}
\toprule
Task & Baseline & Claude & DeepSeek & Gemini & GLM & GPT-5.4 & Grok & Qwen & Seed & GPT-OSS \\
\midrule
Aerodynamics\_CarAerodynamicsSensing  &  0.9617  &  0.9624  &  0.9632  &  0.9632  &  0.9628  &  0.96307  &  0.9624  &  0.9632  &  0.9624  &  0.9625  \\
Astrodynamics\_MannedLunarLanding  &  4577.437  &  6027.3126  &  6079.2455  &  4674.9462  &  6839.0331  &  6660.94  &  4577.437  &  4577.437  &  4733.0435  &  5589.881  \\
ComputerSystems\_MallocLab  &  28  &  96  &  53  &  48  &  86  &  28  &  57  &  32  &  38  &  53  \\
Cryptographic\_AES-128  &  7.5209  &  11.8617  &  12.4591  &  10.2396  &  7.9669  &  39.825  &  10.8615  &  5.5501  &  7.9481  &  12.1639  \\
Cryptographic\_SHA-256  &  9.8274  &  16.7955  &  9.718  &  9.942  &  15.1655  &  26.3405  &  17.2504  &  9.8475  &  15.2838  &  20.4245  \\
Cryptographic\_SHA3-256  &  16.0932  &  17.4003  &  17.0749  &  16.2255  &  17.5778  &  37.4451  &  16.0594  &  16.5292  &  18.3478  &  19.5513  \\
EnergyStorage\_BatteryFastChargingProfile  &  71.2806  &  120.8025  &  111.4518  &  116.6532  &  118.7678  &  121.991  &  99.6875  &  89.8416  &  115.6882  &  113.2709  \\
EnergyStorage\_BatteryFastChargingSPMe  &  66.1636  &  71.8225  &  91.0079  &  92.3198  &  78.0896  &  122.943  &  76.4657  &  79.0273  &  76.4122  &  81.3577  \\
EngDesign  &  1.3571  &  1.3571  &  21.7143  &  27  &  25.5714  &  1.35714  &  27  &  25.5714  &  27  &  22.7857  \\
InventoryOptimization\_disruption\_eoqd  &  0.3642  &  0.6473  &  0.6381  &  0.639  &  0.6303  &  1  &  0.6359  &  0.6225  &  0.6321  &  0.6372  \\
InventoryOptimization\_finite\_horizon\_dp  &  0.3673  &  0.9596  &  0.8025  &  0.7559  &  0.7965  &  0.960684  &  0.8547  &  0.4413  &  0.7323  &  0.7395  \\
InventoryOptimization\_general\_meio  &  0.1825  &  0.9929  &  0.9893  &  0.9839  &  0.9165  &  1  &  0.9236  &  0.7819  &  0.6973  &  0.9542  \\
InventoryOptimization\_joint\_replenishment  &  0.3034  &  0.8822  &  0.8822  &  0.8822  &  0.8822  &  1  &  0.8822  &  0.8821  &  0.8822  &  1  \\
InventoryOptimization\_tree\_gsm\_safety\_stock  &  0.3813  &  0.75  &  0.6606  &  0.6606  &  0.6606  &  1  &  0.6606  &  0.6606  &  0.6606  &  0.6606  \\
JobShop\_abz  &  80.5042  &  96.1035  &  88.3614  &  86.751  &  88.4924  &  91.2314  &  87.6717  &  85.603  &  86.672  &  86.1538  \\
JobShop\_swv  &  81.6325  &  89.4966  &  82.3575  &  82.3141  &  87.1611  &  87.3343  &  85.5068  &  82.6129  &  82.4153  &  86.3069  \\
JobShop\_ta  &  78.8  &  90.8322  &  84.9043  &  85.7065  &  86.8095  &  86.1607  &  84.9136  &  85.5489  &  83.9694  &  83.9922  \\
KernelEngineering\_FlashAttention  &  55.2957  &  983.5001  &  987.2034  &  991.8896  &  381.6257  &  182687.4419  &  324.919  &  525.5567  &  1218.5163  &  716.6822  \\
KernelEngineering\_MLA  &  0.7828  &  1000.3859  &  0.8936  &  1253.2017  &  20.1972  &  1132.0659  &  19.8651  &  0.9271  &  19.987  &  66.7242  \\
KernelEngineering\_TriMul  &  47.1274  &  357.1636  &  85.5923  &  54.5774  &  110.8785  &  47.8829  &  165.0294  &  49.1232  &  84.9069  &  62.7905  \\
Optics\_adaptive\_fault\_tolerant\_fusion  &  0.3959  &  0.6398  &  0.64  &  0.6398  &  0.6398  &  0.455046  &  0.6398  &  0.6398  &  0.6398  &  0.6398  \\
Optics\_adaptive\_temporal\_smooth\_control  &  0.3152  &  0.8419  &  0.8419  &  0.8419  &  0.8417  &  0.84188  &  0.842  &  0.8421  &  0.8421  &  0.8421  \\
Optics\_fiber\_guardband\_spectrum\_packing  &  0.3861  &  0.6692  &  0.657  &  0.6629  &  0.6692  &  0.675429  &  0.6629  &  0.657  &  0.657  &  0.6629  \\
Optics\_fiber\_mcs\_power\_scheduling  &  0.3297  &  0.6542  &  0.5182  &  0.4796  &  0.6491  &  0.660837  &  0.4557  &  0.4458  &  0.6491  &  0.4557  \\
Optics\_fiber\_wdm\_channel\_power\_allocation  &  0.3255  &  0.6675  &  0.6679  &  0.6619  &  0.6686  &  0.696421  &  0.6664  &  0.6666  &  0.6654  &  0.6628  \\
Optics\_holographic\_multifocus\_power\_ratio  &  0.3927  &  0.8072  &  0.8265  &  0.5368  &  0.711  &  1  &  0.4058  &  0.5875  &  0.5626  &  0.8686  \\
Optics\_holographic\_multiplane\_focusing  &  0.3302  &  0.6002  &  0.7196  &  0.4398  &  0.4516  &  1  &  0.474  &  0.5631  &  0.5303  &  0.6757  \\
Optics\_phase\_dammann\_uniform\_orders  &  26.8969  &  99.7995  &  97.3436  &  97.9498  &  97.8709  &  100  &  94.4055  &  95.9998  &  69.0576  &  97.5587  \\
Optics\_phase\_fourier\_pattern\_holography  &  32.6457  &  82.1276  &  74.5838  &  76.6371  &  76.0127  &  100  &  74.217  &  67.3393  &  72.4578  &  74.1263  \\
PyPortfolioOpt\_robust\_mvo\_rebalance  &  32.9804  &  99.9946  &  84.941  &  77.165  &  82.8015  &  99.9946  &  99.983  &  85.5194  &  83.0681  &  99.9946  \\
QuantumComputing\_task\_01\_routing\_qftentangled  &  0.209  &  5.0479  &  3.6155  &  0.209  &  3.7681  &  6.50795  &  3.7655  &  3.2471  &  3.6783  &  0.209  \\
QuantumComputing\_task\_02\_clifford\_t\_synthesis  &  1.7134  &  1.6633  &  1.7134  &  1.7134  &  7.4236  &  1.71337  &  1.6633  &  1.7134  &  1.7134  &  1.6633  \\
QuantumComputing\_task\_03\_cross\_target\_qaoa  &  2.4149  &  2.5781  &  5.103  &  2.9782  &  5.0301  &  2.41491  &  2.6363  &  2.4517  &  2.9782  &  5.2598  \\
ReactionOptimisation\_mit\_case1\_mixed  &  87.3082  &  98.6621  &  98.6041  &  96.5437  &  95.9314  &  98.6621  &  87.3082  &  95.3732  &  95.4297  &  98.6621  \\
ReactionOptimisation\_reizman\_suzuki\_pareto  &  63.5202  &  82.3427  &  82.0329  &  79.473  &  82.9901  &  82.2461  &  63.5202  &  81.4666  &  79.7011  &  100  \\
ReactionOptimisation\_snar\_multiobjective  &  57.5234  &  87.3657  &  82.7881  &  80.1521  &  81.7614  &  100  &  72.3909  &  72.8477  &  79.427  &  74.9518  \\
Robotics\_DynamicObstacleAvoidanceNavigation  &  0.0722  &  0.086  &  0.0856  &  0.0834  &  0.0857  &  0.0857143  &  0.0817  &  0.0765  &  0.0855  &  0.0723  \\
Robotics\_PIDTuning  &  0.0366  &  0.1632  &  0.151  &  0.1521  &  0.1515  &  0.151117  &  0.1585  &  0.1422  &  0.1514  &  0.1454  \\
Robotics\_QuadrupedGaitOptimization  &  0.0218  &  0.0219  &  0.0749  &  0.0218  &  0.1085  &  0.0221543  &  0.0227  &  0.0232  &  0.0218  &  0.33048908  \\
Robotics\_RobotArmCycleTimeOptimization  &  0.2922  &  0.4158  &  0.3923  &  0.4305  &  0.4219  &  0.435621  &  0.3923  &  0.3155  &  0.3256  &  0.3923  \\
Robotics\_UAVInspectionCoverageWithWind  &  28.8519  &  28.8519  &  38.8024  &  28.8519  &  35.1121  &  30.1217  &  55.9109  &  32.8468  &  32.1552  &  43.7417  \\
SingleCellAnalysis\_predict\_modality  &  0.5467  &  0.5467  &  0.5467  &  0.5467  &  0.5467  &  1  &  0.5467  &  0.5467  &  0.5467  &  0.731692725  \\
StructuralOptimization\_ISCSO2015  &  -5401.589  &  -968.4567  &  -1120.212  &  -5401.589  &  -1139.3354  &  -5401.589  &  -1318.7566  &  -1308.2575  &  -1302.2288  &  -1210.9512  \\
StructuralOptimization\_ISCSO2023  &  -77813242.9  &  -16477799.48  &  -55182772.3  &  -20092179.33  &  -17840974.17  &  -77813242.9 &  -30028112.28  &  -66126744.97  &  -42625693.78  &  -48076818.38  \\
StructuralOptimization\_TopologyOptimization  &  -195.9153  &  -190.1498  &  -190.3706  &  -189.3039  &  -188.4673  &  -195.9153  &  -185.7983  &  -192.8488  &  -190.0603  &  -183.0907  \\
SustainableDataCenterControl\_hand\_written\_control  &  8.3294  &  21.5657  &  15.292  &  12.9088  &  19.5978  &  8.5903  &  14.2432  &  30.1873  &  29.2868  &  8.3516  \\
WirelessChannelSimulation\_HighReliableSimulation  &  192.5193  &  292.3228  &  291.9451  &  232.9071  &  248.0119  &  231.224  &  245.7082  &  259.9776  &  304.0437  &  261.7767  \\
\bottomrule
\end{tabular}%
}
\end{table*}

\end{document}